\documentclass{article}

\usepackage[final,nonatbib]{nips_2017}

\usepackage[utf8]{inputenc} 
\usepackage[T1]{fontenc}    
\usepackage{hyperref}       
\usepackage{url}            
\usepackage{booktabs}       
\usepackage{amsfonts}       
\usepackage{nicefrac}       
\usepackage{microtype}      
\usepackage{comment}

\usepackage{wrapfig}
\usepackage[toc,page]{appendix}

\usepackage{subfigure}
\usepackage{amsmath}
\usepackage{amssymb}
\usepackage{bm}
\usepackage{algorithm, algorithmic}

\usepackage{graphicx}

\usepackage[toc,page]{appendix}

\graphicspath{{figures/}}

\def\bfn{{\mathbf{n}}}

\def\bfu{{\mathbf{u}}}
\def\bfv{{\mathbf{v}}}

\def\bfx{{\mathbf{x}}}
\def\bfy{{\mathbf{y}}}
\def\bfz{{\mathbf{z}}}

\def\bfI{{\mathbf{I}}}

\def\calS{{\mathcal{S}}}

\def\calC{{\mathcal{C}}}

\def\calF{{\mathcal{F}}}

\def\calK{{\mathcal{K}}}

\def\calN{{\mathcal{N}}}

\def\calR{{\mathcal{R}}}
\def\calS{{\mathcal{S}}}

\newcommand{\argmin}{\mathrm{arg}\min}

\newcounter{algo}
\renewcommand{\thealgo}{\arabic{algo}}

\title{An Inner-loop Free Solution to Inverse Problems using Deep Neural Networks}

\author{
  Qi Wei$^{\ast}$\\
  Duke University\\
  Durham, NC 27710 \\
  \texttt{qi.wei@duke.edu} \\
  \And
  Kai Fai\thanks{The authors contributed equally to this work.}\\
  Duke University\\
  Durham, NC 27710 \\
  \texttt{kai.fan@stat.duke.edu} \\
  \And
  Lawrence Carin\\
  Duke University\\
  Durham, NC 27710 \\
  \texttt{lcarin@duke.edu} \\  
  \And
  Katherine A. Heller\\
  Duke University\\
  Durham, NC 27710 \\
 \texttt{kheller@stat.duke.edu}
}
\begin{document}

\maketitle

\begin{abstract}
We propose a new method that uses deep learning techniques to accelerate the popular 
alternating direction method of multipliers (ADMM) solution for inverse problems.
The ADMM updates consist of a proximity operator, a least squares regression 
that includes a big matrix inversion, and an explicit solution for updating the dual variables. 
Typically, inner loops are required to solve the first two sub-minimization 
problems due to the intractability of the prior and the matrix inversion.
To avoid such drawbacks or limitations, we propose an \textit{inner-loop free} update 
rule with two pre-trained deep convolutional architectures.  More specifically, we 
learn a conditional denoising auto-encoder which imposes an implicit data-dependent 
prior/regularization on ground-truth in the first sub-minimization problem.
This design follows an empirical Bayesian strategy, leading to so-called amortized inference.
For matrix inversion in the second sub-problem, we learn a convolutional neural
network to approximate the matrix inversion, i.e., the inverse mapping is learned 
by feeding the input through the learned forward network. 
Note that training this neural network does not require ground-truth or measurements,
i.e., it is data-independent. Extensive experiments on both synthetic data and 
real datasets demonstrate the efficiency and accuracy of the proposed method 
compared with the conventional ADMM solution using inner loops
for solving inverse problems.
\end{abstract}

\vspace{-0.5cm}
\section{Introduction}
\vspace{-0.3cm}
Most of the inverse problems are formulated directly to the setting of an optimization
problem related to the a forward model \cite{Tarantola2005inverse}. 
The forward model maps unknown signals, i.e., the ground-truth, 
to acquired information about them, which we call data or measurements. 
This mapping, or forward problem, generally depends on a physical theory that links the ground-truth to the measurements.
Solving inverse problems involves learning the inverse mapping from the measurements to the ground-truth.
Specifically, it recovers a signal from a small number of degraded or noisy measurements.
This is usually ill-posed \cite{Tikhonov1977, Tarantola2005inverse}.
Recently, deep learning techniques have emerged as excellent models and 
gained great popularity for their widespread success in allowing for efficient inference techniques 
on applications include pattern analysis (unsupervised), 
classification (supervised), computer vision, image processing, etc \cite{Deng2014deep}. 
Exploiting deep neural networks to help solve inverse problems has been explored 
recently \cite{sonderby2016amortised, adler2017solving}
and deep learning based methods have achieved state-of-the-art performance in many challenging 
inverse problems like super-resolution \cite{bruna2015super, sonderby2016amortised},
image reconstruction \cite{schlemper2017deep}, automatic colorization \cite{larsson2016learning}.
More specifically, massive datasets currently enables learning end-to-end mappings
from the measurement domain to the target image/signal/data domain 
to help deal with these challenging problems instead of solving the inverse problem 
by inference.
The pairs $\left\{\bfx, \bfy\right\}$ are used to learn the mapping function
from $\bfy$ to $\bfx$ , where $\bfx$ is the ground-truth and $\bfy$ is its corresponding measurement.
This mapping function has recently been characterized by using sophisticated networks, e.g., deep neural networks.
A strong motivation to use neural networks stems from the universal approximation theorem \cite{Csaji2001},
which states that a feed-forward network with a single hidden layer containing a finite number of neurons 
can approximate any continuous function on compact subsets of $\mathbb{R}^{n}$,
under mild assumptions on the activation function.

More specifically, in recent work \cite{bruna2015super, sonderby2016amortised, larsson2016learning,
schlemper2017deep}, an end-to-end mapping from measurements  $\bfy$ to ground-truth $\bfx$
was learned from the training data and then applied to the testing data. Thus, the complicated inference 
scheme needed in the conventional inverse problem solver was replaced by feeding a new measurement through the pre-trained network, which
is much more efficient.  
One main problem for this strategy is that it requires 
task-specific training of the networks, i.e., different problems require different networks. 
Thus, it is very expensive to solve diverse sets of problems. 
To improve the scope of deep neural network models, more recently, in \cite{chang2017one},
a splitting strategy was proposed to decompose an inverse problem
into two optimization problems, where one sub-problem, related to regularization,
can be solved efficiently using trained deep neural networks, 
leading to an alternating direction method of multipliers (ADMM) framework \cite{Boyd2011,7879849}. 
This method involves training a deep convolutional auto-encoder network
for low-level image modeling, which explicitly imposes regularization that spans the subspace 
that the ground-truth images live in. For the sub-problem that requires inverting
a big matrix, a conventional gradient descent algorithm was used, leading to an alternating update,
iterating between feed-forward propagation through a network and iterative gradient descent. 
Thus, an inner loop for gradient descent is still necessary in this framework.


In this work, we propose an inner-loop free framework, in the sense that no iterative algorithm is required to solve sub-problems, 
using a splitting strategy for inverse problems.
The alternating updates for the two sub-problems were derived by feeding 
through two pre-trained deep neural networks, i.e., one using an amortized inference based
denoising convolutional auto-encoder network for the proximity operation and one using structured convolutional
neural networks for the huge matrix inversion related to the forward model.
Thus, the computational complexity of each iteration in ADMM is linear with respect to (w.r.t.) 
the dimensionality of the signals. The network for the proximity operation imposes an implicit prior
learned from the training data, including the measurements  as well as the ground-truth,
leading to amortized inference. The network for matrix inversion is independent from
the training data and can be trained from noise, i.e., a random noise image and its output from
the forward model. This independence from training data allows the proposed framework to 
be used to accelerate almost all the existing training data/example free solutions for inverse
problems based on a splitting strategy. To make training the networks for the proximity operation easier,
three tricks have been employed: the first one is to use a pixel shuffling technique to equalize the 
dimensionality of the measurements and ground-truth; the second one is to optionally add an adversarial
loss borrowed from the GAN (Generative Adversarial Nets) framework \cite{goodfellow2014generative}
for sharp image generation; the last one is to introduce a perceptual measurement loss 
derived from pre-trained networks, such as AlexNet \cite{krizhevsky2012imagenet} 
or VGG-16 Model \cite{simonyan2014very}. 
Arguably, the speed of the proposed algorithm, which we
term Inf-ADMM-ADNN (\emph{Inner-loop free ADMM with Auxiliary Deep Neural Network}),
comes from the fact that it uses two auxiliary pre-trained networks
to accelerate the updates of ADMM.


{\bf{Contribution}} The main contribution of this paper is comprised of
{\bf{i}}) learning an implicit prior/regularizer using a denoising auto-encoder neural network, based on amortized inference;
{\bf{ii}}) learning the inverse of a big matrix using structured convolutional neural networks, without using training data;
{\bf{iii}}) each of the above networks can be exploited to accelerate the existing ADMM solver for inverse problems.

\vspace{-0.4cm}
\section{Linear Inverse Problem}
\vspace{-0.3cm}
\textbf{Notation}: trainable networks by calligraphic font, e.g., $\mathcal{A}$, fixed networks by italic font e.g., $A$. 

As mentioned in the last section,  the low dimensional measurement
is denoted as $\mathbf{y}\in\mathbb{R}^{m}$, which is reduced 
from high dimensional ground truth $\mathbf{x}\in \mathbb{R}^{n}$ 
by a linear operator $A$ such that $\mathbf{y} = A\mathbf{x}$.
Note that usually $n \geq m$, which makes the number of parameters to estimate
no smaller than the number of data points in hand. 
This imposes an ill-posed problem for finding solution $\mathbf{x}$ on new observation $\mathbf{y}$, 
since $A$ is an underdetermined measurement matrix. For example, in a super-resolution set-up, the matrix $A$
might not be invertible, such as the strided Gaussian convolution in \cite{shi2016real, sonderby2016amortised}. 
To overcome this difficulty, several computational strategies, including Markov chain Monte Carlo (MCMC) and 
tailored variable splitting under the ADMM framework,
have been proposed and applied to different kinds of priors, e.g., the empirical Gaussian prior \cite{Wei2015JSTSP,Zhao2016},
the Total Variation prior \cite{Simoes2015,Wei2015FastFusion,Wei2016RFUSE}, etc. 
In this paper, we focus on the popular ADMM framework due to its low computational complexity
and recent success in solving large scale optimization problems. More specifically, the optimization problem is
formulated as 
%
\begin{align}\label{eq:ADMM}
\hat{\mathbf{x}} = \arg\min_{\mathbf{x}, \mathbf{z}} \|\mathbf{y} - A\mathbf{z}\|^2 + \lambda\mathcal{R}(\mathbf{x}), \quad s.t. \quad \mathbf{z} = \mathbf{x}
\end{align}
where the introduced auxiliary variable $\mathbf{z}$ is constrained to be equal to $\bfx$,
and $\mathcal{R}(\mathbf{x})$ captures the structure promoted by the prior/regularization. 
If we design the regularization in an empirical Bayesian way, by imposing an implicit data dependent 
prior on $\mathbf{x}$, i.e., $\mathcal{R}(\mathbf{x}; \mathbf{y})$ for amortized inference \cite{sonderby2016amortised},
the augmented Lagrangian for (\ref{eq:ADMM}) is 
\begin{align}
\mathcal{L}(\mathbf{x}, \mathbf{z}, \mathbf{u}) =  \|\mathbf{y} - A\mathbf{z}\|^2 + \lambda\mathcal{R}(\mathbf{x}; \mathbf{y}) + \langle \mathbf{u}, \mathbf{x} - \mathbf{z} \rangle + \beta \|\mathbf{x} - \mathbf{z}\|^2
\end{align}
where $\mathbf{u}$ is the Lagrange multiplier, and $\beta > 0$ is the penalty parameter. 
The usual augmented Lagrange multiplier method is to minimize $\mathcal{L}$ w.r.t. $\mathbf{x}$ and $\mathbf{z}$ simultaneously. 
This is difficult and does not exploit the fact that the objective function is separable. 
To remedy this issue, ADMM decomposes the minimization into two subproblems that are 
minimizations w.r.t. $\mathbf{x}$ and $\mathbf{z}$, respectively. More specifically, the iterations are as follows: 
\begin{align}
\mathbf{x}^{k+1} &= \arg\min_{\mathbf{x}} \beta \| \mathbf{x} - \mathbf{z}^k + \mathbf{u}^k/2\beta\|^2 + \lambda\mathcal{R}(\mathbf{x}; \mathbf{y}) \label{eq:minx}\\
\mathbf{z}^{k+1} &= \arg\min_{\mathbf{z}} \|\mathbf{y} - A\mathbf{z}\|^2 + \beta \| \mathbf{x}^{k+1} - \mathbf{z} + \mathbf{u}^k /2\beta \|^2 \label{eq:minz}\\
\mathbf{u}^{k+1} &= \mathbf{u}^k + 2\beta(\mathbf{x}^{k+1} - \mathbf{z}^{k+1}). 
\label{eq:updateu}
\end{align}
\vspace{-0.0cm}
If the prior $\mathcal{R}$ is appropriately chosen, such as $\|\mathbf{x}\|_1$, a closed-form solution for $\eqref{eq:minx}$, i.e.,
a soft thresholding solution is naturally desirable. However, for some more complicated regularizations, 
e.g., a patch based prior \cite{Elad2006}, solving \eqref{eq:minx} is nontrivial, and may require iterative methods.
To solve \eqref{eq:minz}, a matrix inversion is necessary, for which conjugate gradient descent (CG)
is usually applied to update $\mathbf{z}$ \cite{chang2017one}. 
Thus, solving \eqref{eq:minx} and \eqref{eq:minz} is in general cumbersome. 
Inner loops are required to solve these two sub-minimization problems due to the intractability of the prior and the inversion, 
resulting in large computational complexity.
To avoid such drawbacks or limitations, we propose an \textit{inner loop-free} update rule with two pretrained deep convolutional architectures. 

\vspace{-0.1cm}
\section{Inner-loop free ADMM}
\vspace{-0.1cm}
\subsection{Amortized inference for $\mathbf{x}$ using a conditional proximity operator}

Solving sub-problem (\ref{eq:minx}) is equivalent to finding the solution of the proximity operator 
\begin{align}
\mathcal{P}_{\mathcal{R}}(\mathbf{v};\mathbf{y})=\arg\min_{\mathbf{x}} \frac{1}{2}\|\mathbf{x} - \mathbf{v}\|^2 + \mathcal{R}(\mathbf{x};\mathbf{y})
\end{align}
where we incorporate the constant $\frac{\lambda}{2\beta}$ into $\mathcal{R}$ without loss of generality. 
If we impose the first order necessary conditions \cite{maurer1979first}, we have
\begin{align}\label{eq:1stcond}
\mathbf{x}=\mathcal{P}_{\mathcal{R}}(\mathbf{v};\mathbf{y}) \Leftrightarrow 0 \in \partial \mathcal{R}(\cdot;\mathbf{y})(\mathbf{x}) + \mathbf{x} - \mathbf{v} \Leftrightarrow \mathbf{v} - \mathbf{x} \in \partial \mathcal{R}(\cdot;\mathbf{y})(\mathbf{x})
\end{align}
where $\partial \mathcal{R}(\cdot;\mathbf{y})$ is a partial derivative operator. 
For notational simplicity, we define another operator $\mathcal{F} =: \mathcal{I} + \partial \mathcal{R}(\cdot;\mathbf{y})$.
Thus, the last condition in (\ref{eq:1stcond}) indicates that $\mathbf{x}^{k+1}=\mathcal{F}^{-1}(\mathbf{v})$. 
Note that the inverse here represents the inverse of an operator, i.e., the inverse function of $\calF$. 
Thus our objective is to learn such an inverse operator which projects $\mathbf{v}$ into the prior subspace. 
For simple priors like $\|\cdot\|_1$ or $\|\cdot\|_2^2$, the projection can be efficiently computed. 
In this work, we propose an implicit example-based prior, which does not have a truly Bayesian interpretation, 
but aids in model optimization. In line with this prior, we define the 
implicit proximity operator $\mathcal{G}_{\bm\theta}(\mathbf{x}; \mathbf{v}, \mathbf{y})$ 
parameterized by $\bm\theta$ to approximate unknown $\mathcal{F}^{-1}$. 
More specifically, we propose a neural network architecture referred to as conditional 
Pixel Shuffling Denoising Auto-Encoders (cPSDAE) as the operator $\mathcal{G}$, 
where pixel shuffling \cite{shi2016real} means periodically reordering the pixels 
in each channel mapping a high resolution image to a low resolution image with scale $r$ and 
increase the number of channels to $r^2$ (see \cite{shi2016real} for more details). 
This allows us to transform $\mathbf{v}$ so that it is the same scale as $\mathbf{y}$,
and concatenate it with $\bfy$ as the input of cPSDAE easily. The architecture of cPSDAE is shown in Fig. \ref{fig:znet} (d).

\vspace{-0.2cm}
\subsection{Inversion-free update of $\mathbf{z}$}
\begin{figure}
\centering
    \subfigure[]{
    \includegraphics[width=0.24\textwidth]{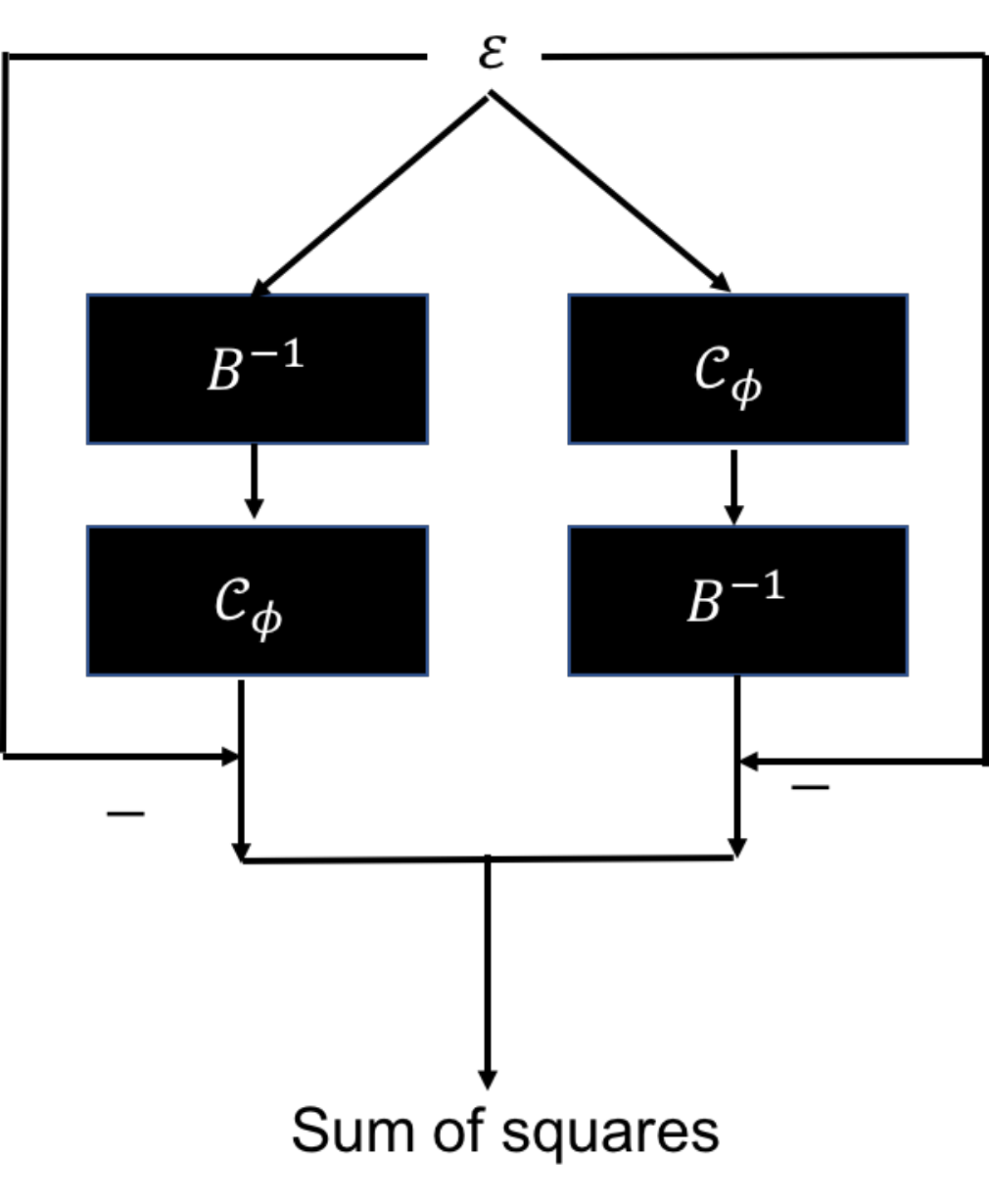}}
    \subfigure[]{
    \includegraphics[width=0.12\textwidth]{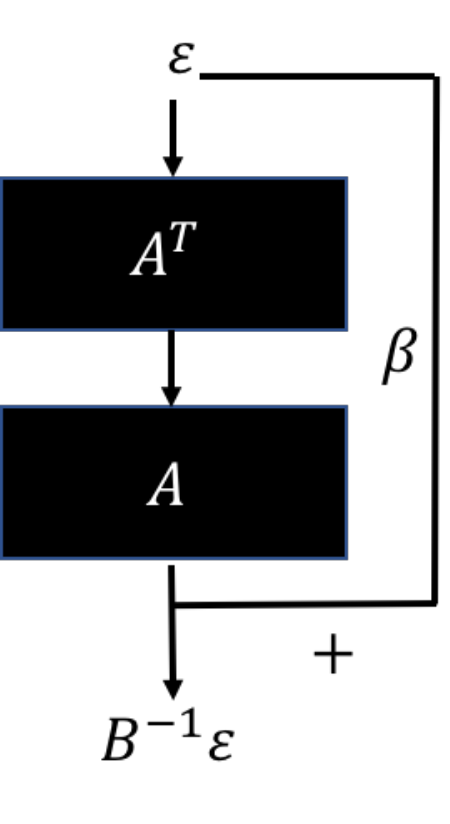}}
	\subfigure[]{
    \includegraphics[width=0.1\textwidth]{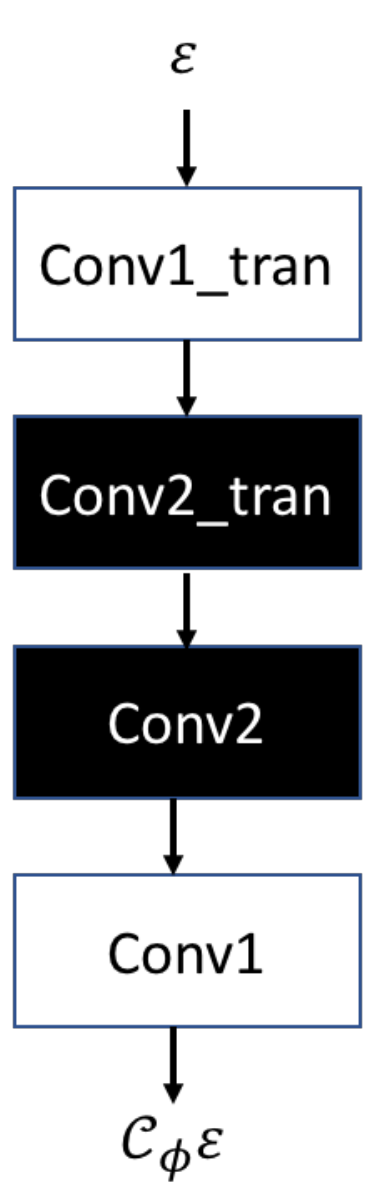}}
    \subfigure[]{
    \includegraphics[width=0.24\textwidth]{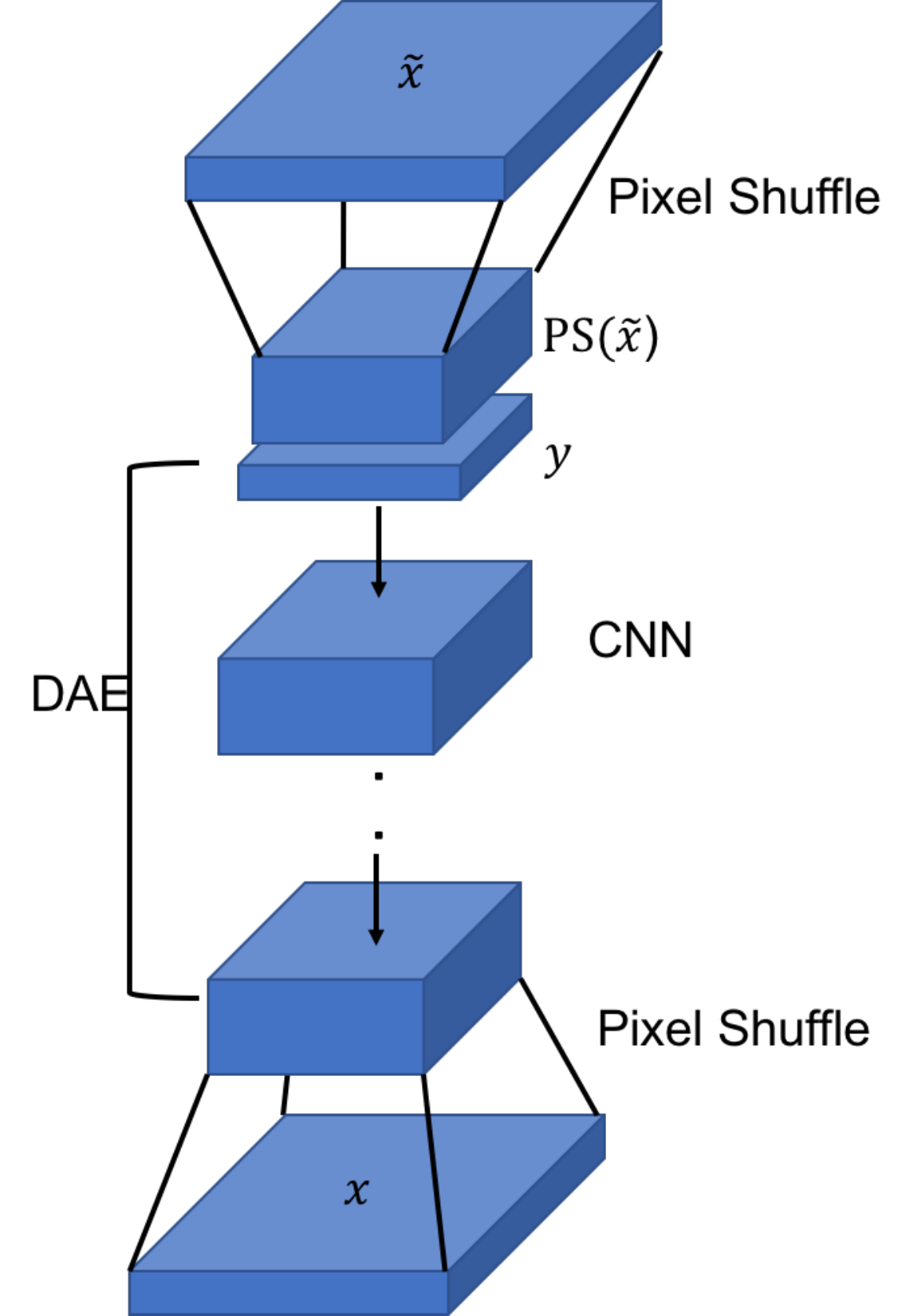}}
    \subfigure[]{
    \includegraphics[width=0.24\textwidth]{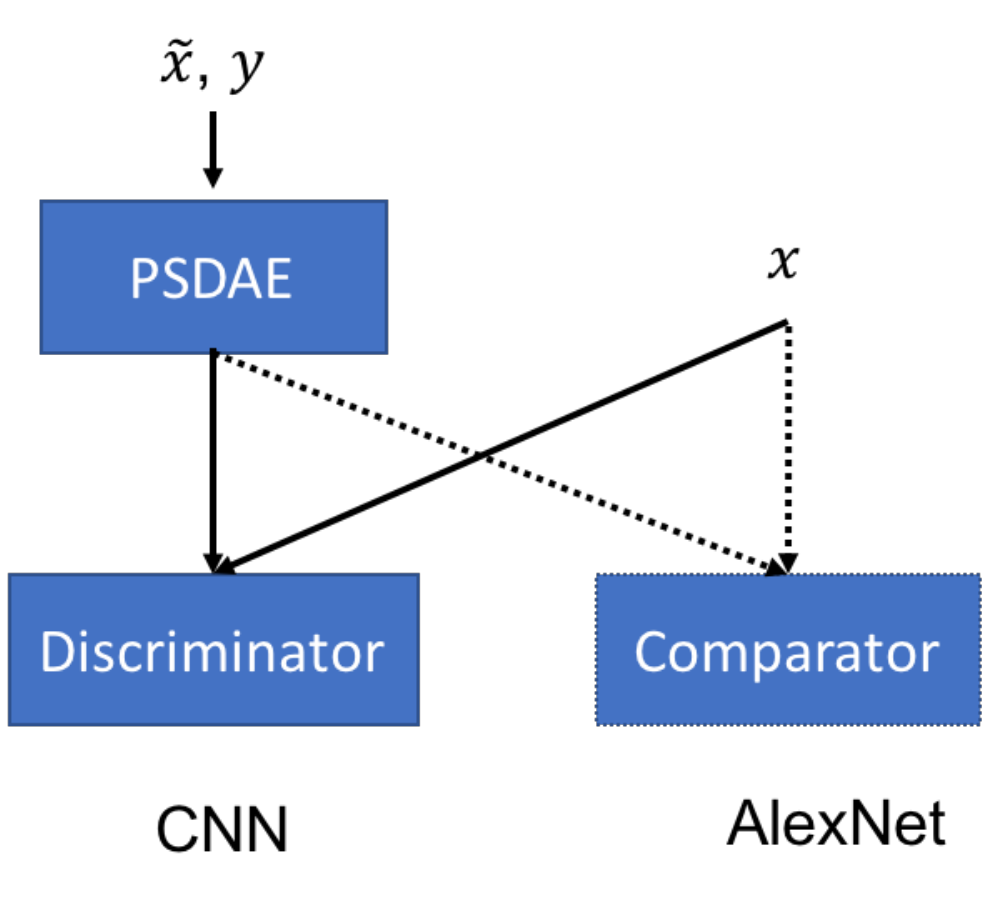}}
    \caption{\footnotesize{Network for updating $\bfz$ (in black): (a) loss function \eqref{eq:loss_z}, (b) structure of $B^{-1}$, (c) struture of $\cal{C}_{\bm \phi}$. 
   	     		   Note that the input $\epsilon$ is random noise independent from the training data. Network for updating $\bfz$ (in blue): (d) structure of cPSDAE $\mathcal{G}_{\bm\theta}(\mathbf{x}; \tilde{\bfx}, \mathbf{y})$  ($\tilde{\bfx}$ plays the same role as $\bfv$ in training), (e) adversarial training for $ \mathcal{R}(\mathbf{x};\mathbf{y})$. Note again that (a)(b)(c) describes the network for inferring $\bfz$, which is data-independent  and (d)(e) describes the network for inferring $\bfx$, which is data-dependent.}}
\label{fig:znet}
\end{figure}

While it is straightforward to write down the closed-form solution for sub-problem (\ref{eq:minz}) w.r.t. $\bfz$ as is shown in \eqref{eq:updatez},
explicitly computing this solution is nontrivial.
\begin{align}\label{eq:updatez}
\mathbf{z}^{k+1} = K \left(A^\top \mathbf{y} + \beta\mathbf{x}^{k+1} + \mathbf{u}^k/2  \right), \text{ where } K=\left(A^\top A + \beta \mathbf{I} \right)^{-1}
\end{align}
In \eqref{eq:updatez}, $A^\top$ is the transpose of the matrix $A$. 
As we mentioned, the term $K$ in the right hand side involves an expensive matrix inversion with computational complexity $O(n^3)$ .
Under some specific assumptions, e.g., $A$ is a circulant matrix, this matrix inversion can be accelerated
with a Fast Fourier transformation, which has a complexity of order $\mathcal{O}(n\log n)$. 
Usually, the gradient based update has linear complexity in each iteration and thus has an overall complexity of 
order $\mathcal{O}( n_{\textrm{int}} \log n)$, where $n_{\textrm{int}}$ is the number of iterations.
In this work, we will learn this matrix inversion explicitly by
designing a neural network. Note that $K$ is only dependent on $A$, and thus can be computed in advance 
for future use. This problem can be reduced to a smaller scale matrix inversion by applying the Sherman-Morrison-Woodbury formula:
\begin{align}
K = \beta^{-1} \left(\mathbf{I} - A^\top B A\right), \text{ where } B = \left( \beta \mathbf{I} + AA^\top \right)^{-1}.
\label{eq:K_eq}
\end{align}
Therefore, we only need to solve the matrix inversion in dimension $m \times m$, i.e., estimating $B$. 
We propose an approach to approximate it by a trainable deep convolutional neural network $\mathcal{C}_{\bm\phi} \approx B$ 
parameterized by $\bm\phi$. Note that $B^{-1}=\lambda \mathbf{I} + AA^\top$ can be considered 
as a two-layer fully-connected or convolutional network as well, but with a fixed kernel. 
This inspires us to design two auto-encoders with shared weights, and minimize the sum of two reconstruction losses
to learn the inversion $\mathcal{C}_{\bm\phi}$ :
\begin{align}
\arg\min_{\bm\phi} \mathbb{E}_{\bm \varepsilon}\left[ \|  {\bm \varepsilon} - \mathcal{C}_{\bm\phi} B^{-1}{\bm{\varepsilon}}\|_2^2 + \| {\bm{\varepsilon}} -  B^{-1} \mathcal{C}_{\bm\phi} {\bm{\varepsilon}}\|_2^2 \right]
\label{eq:loss_z}
\end{align}
where $\bm{\varepsilon}$ is sampled from a standard Gaussian distribution.
The loss in \eqref{eq:loss_z} is clearly depicted in Fig. \ref{fig:znet} (a)
with the structure of $B^{-1}$ in Fig. \ref{fig:znet} (b) and the structure
of $\calC_{\bm\phi}$ in Fig. \ref{fig:znet} (c).
Since the matrix $B$ is symmetric, we can reparameterize $\mathcal{C}_{\bm\phi}$ as $\mathcal{W}_{\bm\phi}\mathcal{W}_{\bm\phi}^\top$,
where $\mathcal{W}_{\bm\phi}$ represents a multi-layer convolutional network 
and $\mathcal{W}_{\bm\phi}^\top$ is a symmetric 
convolution transpose architecture using shared kernels with $\mathcal{W}_{\bm\phi}$,
as shown in Fig. \ref{fig:znet} (c) (the blocks with the same colors share the same network parameters). 
By plugging the learned $\mathcal{C}_{\bm\phi}$ in \eqref{eq:K_eq} , we obtain a reusable deep 
neural network $\mathcal{K}_{\bm\phi}=\beta^{-1} \left(\mathbf{I} - A^\top \mathcal{C}_{\bm\phi} A\right)$
as a surrogate for the exact inverse matrix $K$. The update of $\mathbf{z}$ at each iteration can be done 
by applying the same $\mathcal{K}_{\bm\phi}$ as follows:
\begin{align}\label{eq:updatez_all}
\mathbf{z}^{k+1} \leftarrow \beta^{-1} \left(\mathbf{I} - A^\top  \mathcal{C}_{\bm\phi} A\right) \left(A^\top \mathbf{y} + \beta\mathbf{x}^{k+1} + \mathbf{u}^k/2  \right).
\end{align}

\vspace{-0.2cm}

\vspace{-0.3cm}
\subsection{Adversarial training of cPSDAE}
In this section, we will describe the proposed adversarial training scheme for cPSDAE to update $\mathbf{x}$.
Suppose that we have the paired training dataset $(\mathbf{x}_i, \mathbf{y}_i)_{i=1}^N$, a single cPSDAE with the
input pair $(\tilde{\mathbf{x}}, \mathbf{y})$ is trying to minimize the reconstruction error
$\mathcal{L}_r(\mathcal{G}_{\bm\theta}(\tilde{\mathbf{x}},\mathbf{y}), \mathbf{x})$,
where $\tilde{\mathbf{x}}$ is a corrupted version of $\mathbf{x}$, i.e., $\tilde{\bfx} = \bfx + \bfn$ where $\bfn$ is random noise. 
Notice $\mathcal{L}_r$ in traditional DAE is commonly defined as $\ell_2$ loss, however, $\ell_1$ loss is an alternative in practice. 
Additionally, we follow the idea in \cite{nguyen2016plug, dosovitskiy2016generating} by introducing a discriminator and a 
comparator to help train the cPSDAE, and find that it can produce sharper or higher quality images 
than merely optimizing $\mathcal{G}$. This will wrap our conditional generative model $\mathcal{G}_{\bm\theta}$ into the
conditional GAN \cite{goodfellow2014generative} framework with an extra feature matching network (comparator). 
Recent advances in representation learning problems have shown that the features extracted from well pre-trained 
neural networks on supervised classification problems can be successfully transferred to others tasks, 
such as zero-shot learning \cite{lei2015predicting}, style transfer learning \cite{gatys2016image}. 
Thus, we can simply use pre-trained AlexNet \cite{krizhevsky2012imagenet} or VGG-16 Model \cite{simonyan2014very} 
on ImageNet as the comparator without fine-tuning in order to extract features that capture complex and perceptually
important properties. The feature matching loss $\mathcal{L}_f(C(\mathcal{G}_{\bm\theta}(\tilde{\mathbf{x}}, \mathbf{y})), C(\mathbf{x}))$ is usually the $\ell_2$ distance of high level image features, where $C$ represents the pre-trained network.
Since $C$ is fixed, the gradient of this loss can be back-propagated to $\bm\theta$. 

For the adversarial training, the discriminator $\mathcal{D}_{\bm\psi}$ is a trainable convolutional network. 
We can keep the standard discriminator loss as in a traditional GAN, and add the generator loss of the GAN 
to the previously defined DAE loss and comparator loss. 
Thus, we can write down our two objectives as follows,
\begin{align}
\mathcal{L}_D(\mathbf{x}, \mathbf{y}) &= - \log \mathcal{D}_{\bm\psi}(\mathbf{x}) - \log \left( 1 - \mathcal{D}_{\bm\psi}(\mathcal{G}_{\bm\theta}(\tilde{\mathbf{x}}, \mathbf{y})) \right) \\
\mathcal{L}_G(\mathbf{x}, \mathbf{y}) &= \lambda_r\|\mathcal{G}_{\bm\theta}(\tilde{\mathbf{x}}, \mathbf{y}) - \mathbf{x}\|_2^2 + \lambda_f\|C(\mathcal{G}_{\bm\theta}(\tilde{\mathbf{x}}, \mathbf{y})) - C(\mathbf{x})\|_2^2 - \lambda_a \log \mathcal{D}_{\bm\psi}(\mathcal{G}_{\bm\theta}(\tilde{\mathbf{x}}, \mathbf{y})) 
\end{align}
The optimization involves iteratively updating $\bm\psi$ by minimizing $\mathcal{L}_D$ keeping $\bm\theta$ fixed, and then updating $\bm\theta$ by minimizing $\mathcal{L}_G$ keeping $\bm\psi$ fixed. 
The proposed method, including training and inference has been summarized in Algorithm \ref{alg:InfADMM}. Note that each update of $\bfx$ or $\bfz$
using neural networks in an ADMM iteration has a complexity of linear order w.r.t. the data dimensionality $n$.

\begin{wrapfigure}{L}{0.5\textwidth}
	\begin{minipage}{0.5\textwidth}
	\vspace{-0.8cm}
		\begin{algorithm}[H]
		\caption{Inner-loop free ADMM with Auxiliary Deep Neural Nets (Inf-ADMM-ADNN)}
		\label{alg:InfADMM}
		\textit{Training stage}: 
		\begin{algorithmic}[1]
		
		\STATE Train net $\calK_{\bm\phi}$ for inverting $A^TA+\beta \bfI$
		\STATE Train net cPSDAE for proximity operator of $\calR(\bfx; \bfy)$ 
		\end{algorithmic}
		
		\textit{Testing stage}: 
		\begin{algorithmic}[1]
		%
		\FOR{$ t = 1,2,\dots $}
		\STATE Update $\bfx$ cf. $\bfx^{k+1} = \calF^{-1} (\bfv)$;
		\STATE Update $\bfz$ cf. \eqref{eq:updatez_all};
		\STATE Update $\bfu$ cf. \eqref{eq:updateu};
		\ENDFOR
		\end{algorithmic}
		\end{algorithm}
	\end{minipage}
\end{wrapfigure}
\subsection{Discussion}
A critical point for learning-based methods is whether the method generalizes to other problems.
More specifically, how does a method that is trained on a specific dataset perform when applied
to another dataset? To what extent can we reuse the trained network without re-training?

In the proposed method, two deep neural networks are trained to infer $\bfx$ and $\bfz$.
For the network w.r.t. $\bfz$, the training only requires the forward model $A$ to generate 
the training pairs ($\epsilon , A \epsilon$). The trained network for $\bfz$
can be applied for any other datasets as long as they share the same $A$.
Thus, this network can be adapted easily to accelerate inference for inverse problems 
without training data. However, for inverse problems that depends on a different $A$,
a re-trained network is required. It is worth mentioning that the forward model
$A$ can be easily learned using training dataset $(\bfx, \bfy)$, leading to a 
fully blind estimator associated with the inverse problem. An example of learning
$\hat{A}$ can be found in the supplementary materials (see Section 1).
For the network w.r.t. $\bfx$, training requires data pairs $(\bfx_i, \bfy_i)$ because of the
amortized inference. Note that this is different from training a prior for $\bfx$ only 
using training data $\bfx_i$. Thus, the trained network for $\bfx$ is confined to the specific tasks
constrained by the pairs ($\bfx, \bfy$). To extend the generality of the trained network,
the amortized setting can be removed, i.e, the measurements $\bfy$ is removed 
from the training, leading to a solution to proximity operator $\mathcal{P}_{\mathcal{R}}(\mathbf{v})=\arg\min_{\mathbf{x}} \frac{1}{2}\|\mathbf{x} - \mathbf{v}\|^2 + \mathcal{R(\bfx)}$. This proximity operation can be regarded as a denoiser which
projects the noisy version $\bfv$ of $\bfx$ into the subspace imposed by $\calR(\bfx)$.
The trained network (for the proximity operator) can be used as a plug-and-play prior \cite{venkatakrishnan2013plug} to
regularize other inverse problems for datasets that share similar
statistical characteristics. However, a significant change in the training dataset, e.g.,
different modalities like MRI and natural images (e.g., ImageNet \cite{krizhevsky2012imagenet}), 
would require re-training. 

Another interesting point to mention is the scalability of the proposed method to data
of different dimensions. The scalability can be adapted using patch-based methods without loss of generality.
For example, a neural network is trained for images of size $64 \times 64$ but the test image is of 
size $256 \times 256$. To use this pre-trained network, the full image can be decomposed as 
four $64 \times 64$ images and fed to the network. To overcome the possible blocking artifacts,
eight overlapping patches can be drawn from the full image and fed to the network.
The output of these eight patches are then averaged (unweighted or weighted) over the overlapping parts.
A similar strategy using patch stitching can be exploited to feed small patches 
to the network for higher dimensional datasets.


\vspace{-0.3cm}
\section{Experiments}
\vspace{-0.3cm}
In this section, we provide experimental results and analysis on the proposed Inf-ADMM-ADNN
and compare the results with a conventional ADMM using inner loops for inverse problems.
Experiments on synthetic data have been implemented to show 
the fast convergence of our method, which comes from the efficient feed-forward propagation 
through pre-trained neural networks.  Real applications using proposed
Inf-ADMM-ADNN have been explored, including 
single image super-resolution, motion deblurring and joint super-resolution and colorization. 
\vspace{-0.1cm}
\subsection{Synthetic data}
\begin{figure}[H]
\centering
\includegraphics[width=0.32\textwidth]{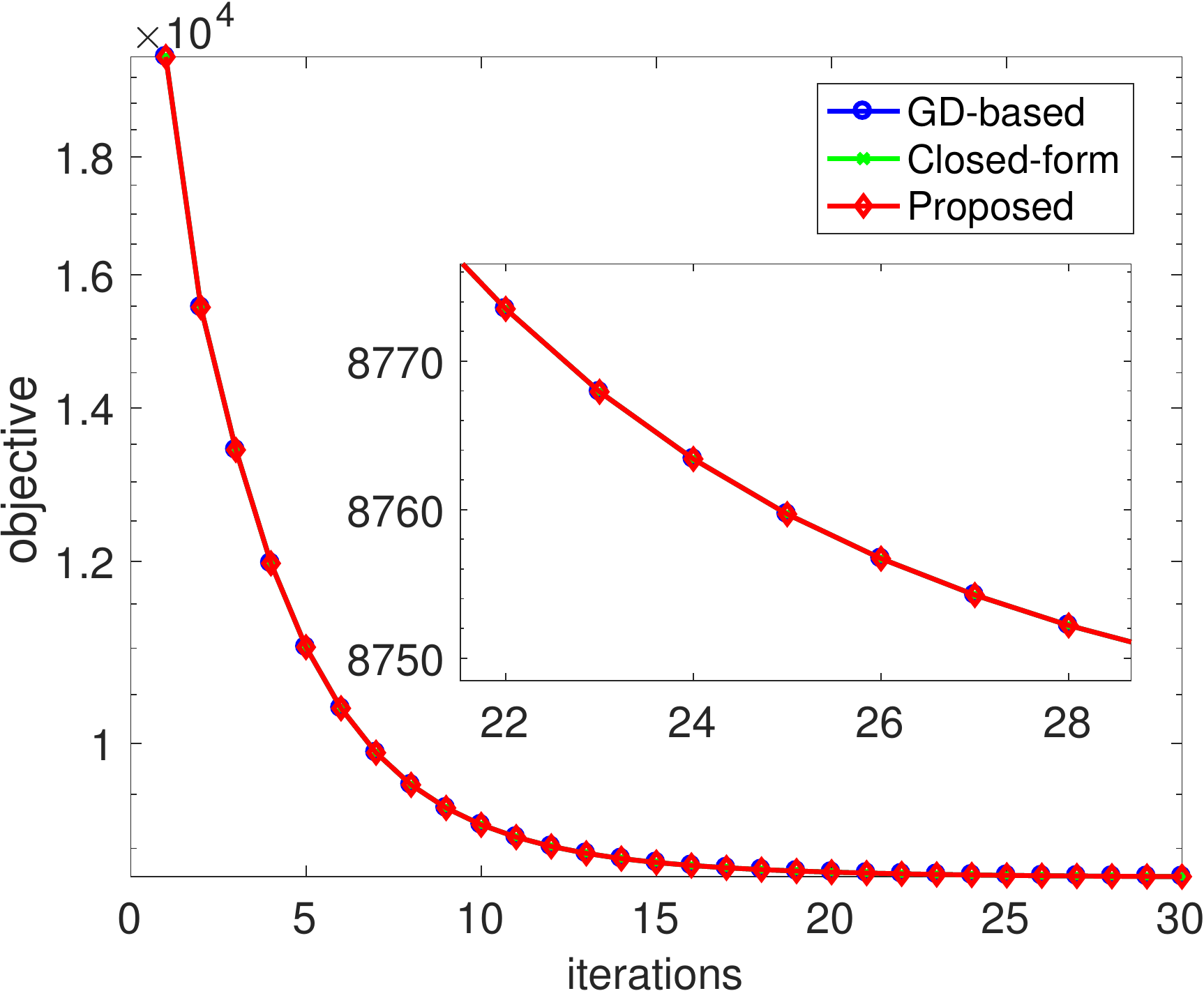}
\includegraphics[width=0.32\textwidth]{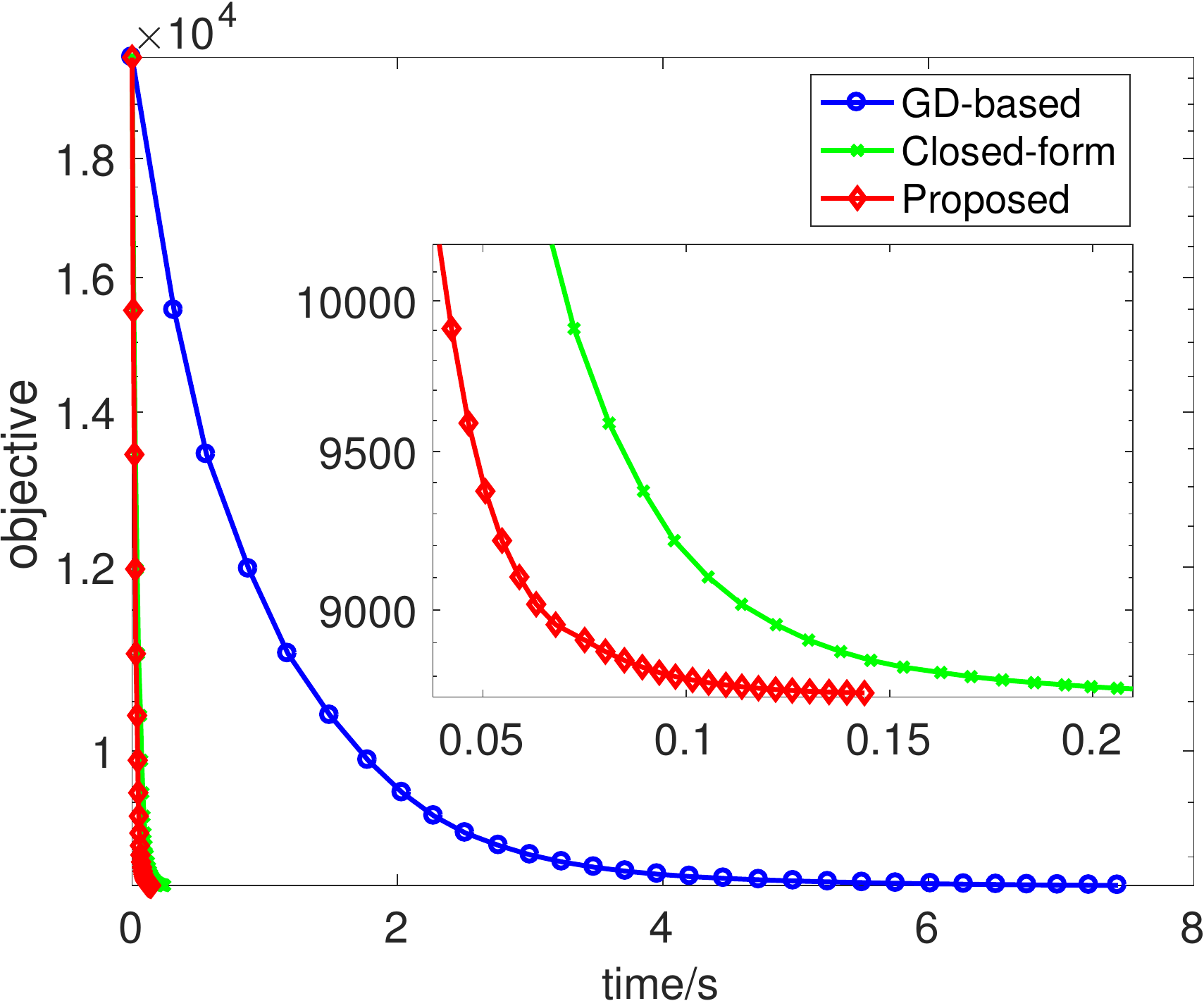}
\includegraphics[width=0.32\textwidth]{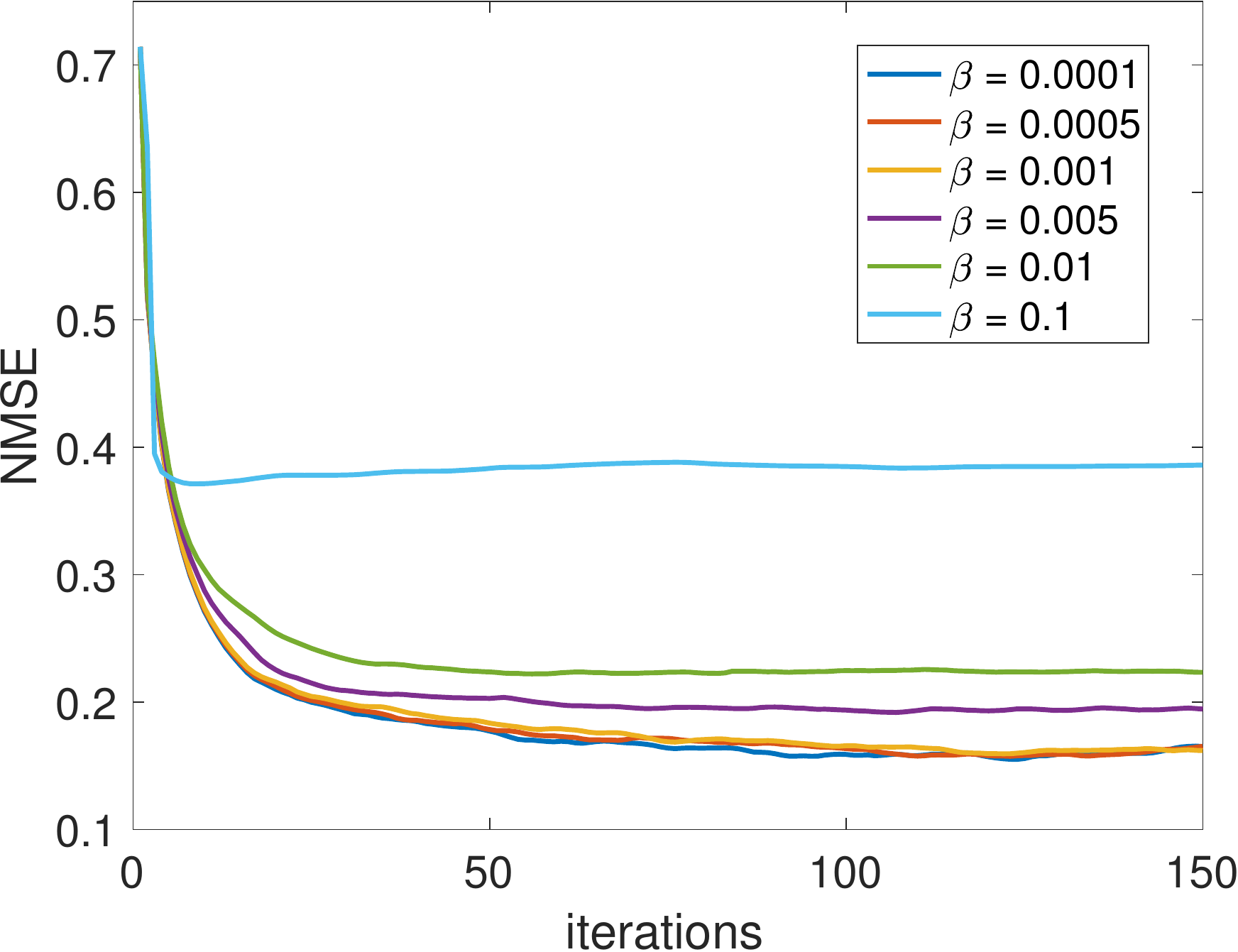}
\caption{\footnotesize{Synthetic data: (left) objective \emph{v.s.} iterations, (middle) objective \emph{v.s.} time.
MNIST dataset: (right) NMSE \emph{v.s.} iterations for MNIST image $4 \times$ super-resolution.}}
\label{fig:obj_iter}
\end{figure}

\vspace{-0.1cm}
To evaluate the performance of proposed Inf-ADMM-ADNN, we first 
test the neural network $\calK_{\bm\phi}$, approximating the matrix 
inversion on synthetic data. More specifically, we assume that the ground-truth $\bfx$ 
is drawn from a Laplace distribution $\textrm{Laplace}(\mu,b)$, where $\mu=0$ is the
location parameter and $b$ is the scale parameter. The forward model $A$ 
is a sparse matrix representing convolution with a stride of $4$.
The architecture of $A$ is available in the supplementary materials (see Section 2). 
The noise $\bfn$ is drawn from a standard Gaussian distribution $\calN(0,\sigma^2)$.
Thus, the observed data is generated as $\bfy = A \bfx +\bfn$.
Following Bayes theorem, the maximum a posterior estimate of $\bfx$ given $\bfy$, i.e., 
maximizing $p(\bfx|\bfy) \varpropto  p(\bfy|\bfx) p(\bfx)$ can be equivalently formulated as 
$\argmin_{\bfx} \frac{1}{2 \sigma^2}\|\bfy-A \bfx\|_2^2 + \frac{1}{b} \|\bfx\|_1$,
where $b=1$ and $\sigma=1$ in this setting.
Following \eqref{eq:minx}, \eqref{eq:minz}, \eqref{eq:updateu}, this problem is reduced to
the following three sub-problems:
\vspace{-0.1cm}
\begin{align}
\mathbf{x}^{k+1} &= \calS_{\frac{1}{2\beta}}(\bfz^{k} - \bfu^{k} / 2\beta)\\
\mathbf{z}^{k+1} &= \arg\min_{\mathbf{z}} \|\mathbf{y} - A\mathbf{z}\|_2^2 + \beta \| \mathbf{x}^{k+1} - \mathbf{z} + \mathbf{u}^k /2\beta \|_2^2 \\
\mathbf{u}^{k+1} &= \mathbf{u}^k + 2\beta(\mathbf{x}^{k+1} - \mathbf{z}^{k+1})
\end{align}
\vspace{-0.1cm}
where the soft thresholding operator $\calS$ is defined as 
$\calS_{\kappa}(a) = \left\{\begin{array}{ccc}
& 0  & |a|  \leq \kappa \\
& a - \textrm{sgn}(a)\kappa & |a| > \kappa 
\end{array}
\right.
$
and sgn($a$) extracts the sign of $a$.
The update of $\bfx^{k+1}$ has a closed-form solution, i.e., soft thresholding of $\bfz^{k}-\bfu^{k}/{2 \beta}$.
The update of $\bfz^{k+1}$ requires the inversion of a big matrix, which is usually solved using a gradient descent based algorithm.
The update of $\bfu^{k+1}$ is straightforward. Thus, we compare the gradient descent based update, 
a closed-form solution for matrix inversion\footnote{Note that this matrix inversion
can be explicitly computed due to its small size in this toy experiment. In practice, this matrix is not built explicitly.} 
and the proposed inner-free update using a pre-trained neural network.
The evolution of the objective function w.r.t. the number of iterations and the time has been plotted in the left and middle of Figs. \ref{fig:obj_iter}.
While all three methods perform similarly from iteration to iteration (in the left of Figs. \ref{fig:obj_iter}),
the proposed inner-loop free based and closed-form inversion based methods converge much faster than the 
gradient based method (in the middle of Figs. \ref{fig:obj_iter}).
Considering the fact that the closed-form solution, i.e., a direct matrix inversion, is usually not available in practice, 
the learned neural network allows us to approximate the matrix inversion in a very accurate and efficient way.
\vspace{-0.3cm}

%


\begin{figure}
\centering
\includegraphics[width=0.24\textwidth]{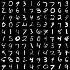}
\includegraphics[width=0.24\textwidth]{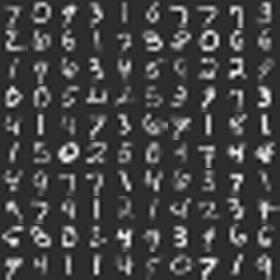}
\includegraphics[width=0.24\textwidth]{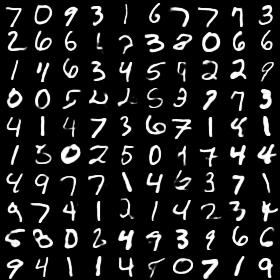}
\includegraphics[width=0.24\textwidth]{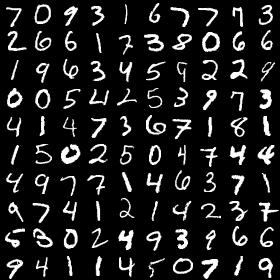}
\includegraphics[width=0.24\textwidth]{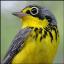}
\includegraphics[width=0.24\textwidth]{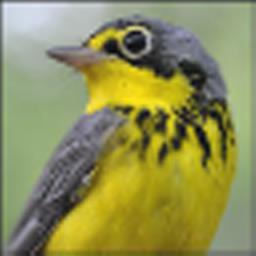}
\includegraphics[width=0.24\textwidth]{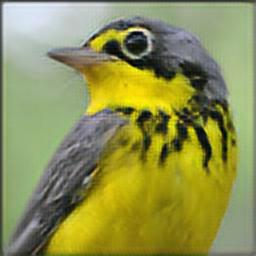}
\includegraphics[width=0.24\textwidth]{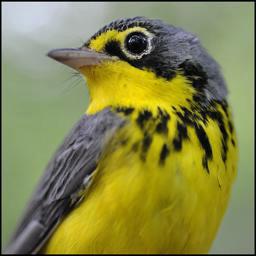}
\includegraphics[width=0.24\textwidth]{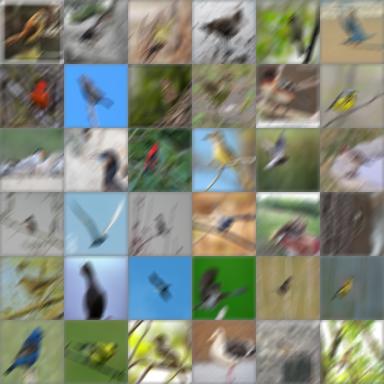}
\includegraphics[width=0.24\textwidth]{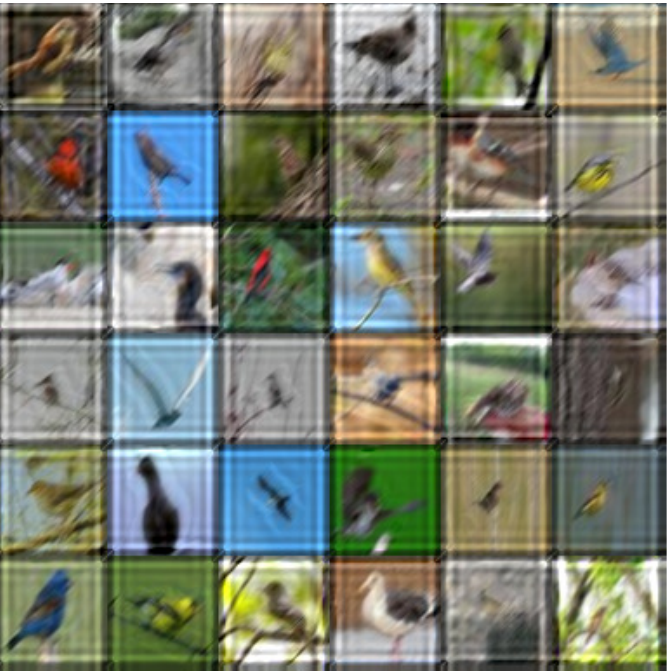} 
\includegraphics[width=0.24\textwidth]{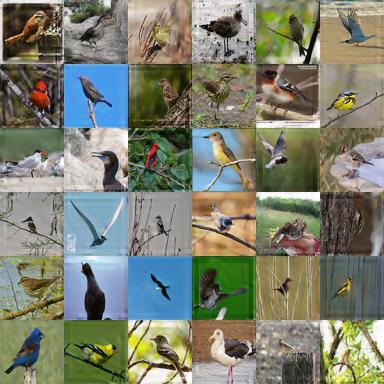}
\includegraphics[width=0.24\textwidth]{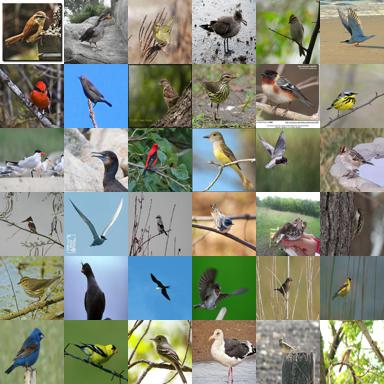} 
\caption{\footnotesize{Top two rows : (column 1) LR images, (column 2) bicubic interpolation ($\times 4$), (column 3) results using proposed method ($\times 4$), (column 4) HR image. Bottom row: (column 1) motion blurred images, (column 2) results using Wiener filter with the best performance by tuning regularization parameter, (column 3) results using proposed method, (column 4) ground-truth.}}
\label{fig:SR_all}
\end{figure}

\subsection{Image super-resolution and motion deblurring}

In this section, we apply the proposed Inf-ADMM-ADNN to solve the poplar image 
super-resolution problem. We have tested our algorithm on the MNIST 
dataset \cite{lecun1998gradient} and the $11$K images of the 
Caltech-UCSD Birds-200-2011 (CUB-200-2011) dataset \cite{wah2011caltech}.
In the first two rows of Fig. \ref{fig:SR_all}, high resolution images, as shown in the last column,
have been blurred (convolved) using a Gaussian kernel of size $3 \times 3$ and downsampled
every $4$ pixels in both vertical and horizontal directions to generate the corresponding low 
resolution images as shown in the first column. The bicubic interpolation of LR images and results using proposed Inf-ADMM-ADNN on a $20\%$
held-out test set are displayed in column $2$ and $3$. Visually, the proposed Inf-ADMM-ADNN gives
much better results than the bicubic interpolation, recovering more details including colors and edges.
A similar task to super-resolution is motion deblurring, in which the convolution kernel is a 
directional kernel and there is no downsampling. The motion deblurring results using Inf-ADMM-ADNN 
are displayed in the bottom of Fig. \ref{fig:SR_all} and are compared with the Wiener filtered deblurring
result (the performance of Wiener filter has been tuned to the best by adjusting the regularization parameter).
Obviously, the Inf-ADMM-ADNN gives visually much better results than the Wiener filter.
Due to space limitations, more simulation results are available in supplementary 
materials (see Section 3.1 and 3.2).

To explore the convergence speed w.r.t. the ADMM 
regularization parameter $\beta$, we have plotted the normalized mean square error (NMSE) 
defined as $\textrm{NMSE} = {\|\hat{\bfx}-\bfx\|_2^2}/{\|\bfx\|_2^2}$, 
of super-resolved MNIST images w.r.t. ADMM iterations using different values of $\beta$
in the right of Fig. \ref{fig:obj_iter}. It is interesting to note that when $\beta$ is large, e.g., $0.1$ or $0.01$,
the NMSE of ADMM updates converges to a stable value rapidly in a few iterations (less than $10$).
Reducing the value of $\beta$ slows down the decay of NMSE over iterations but reaches 
a lower stable value. When the value of $\beta$ is small enough, e.g., $\beta=0.0001, 0.0005, 0.001$,
the NMSE converges to the identical value. This fits well with the claim in Boyd's book \cite{Boyd2011} that 
when $\beta$ is too large it does not put enough emphasis on minimizing the objective function, causing
coarser estimation; thus a relatively small $\beta$ is encouraged in practice. Note that the selection
of this regularization parameter is still an open problem.

\subsection{Joint super-resolution and colorization}
\begin{figure}[H]
\includegraphics[width=0.325\textwidth]{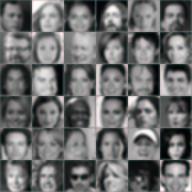} 
\includegraphics[width=0.325\textwidth]{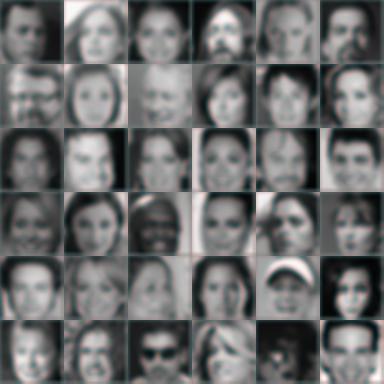}
\includegraphics[width=0.325\textwidth]{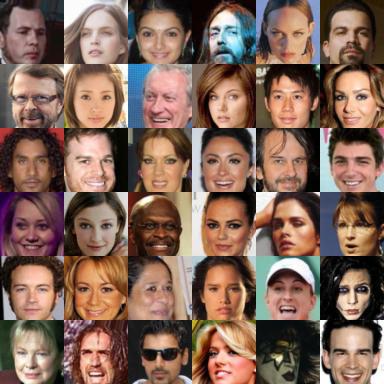}\\
\includegraphics[width=0.325\textwidth]{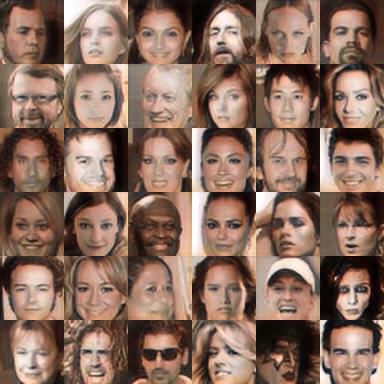} 
\includegraphics[width=0.325\textwidth]{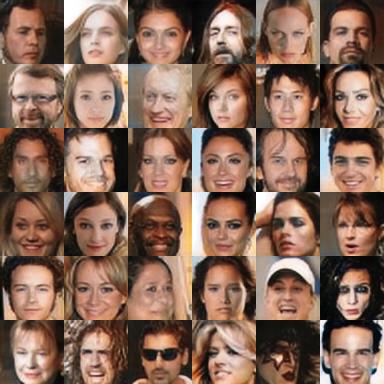} 
\includegraphics[width=0.325\textwidth]{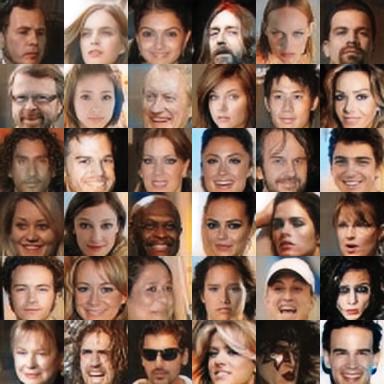} 
\caption{\footnotesize{(top left) colorless LR image, (top middle) bicubic interpolation, (top right) HR ground-truth, 
(bottom left to right) updated image in $\mathbf{1}$th, $\mathbf{4}$th and $\mathbf{7}$th ADMM iteration. 
Note that the colorless LR images and bicubic interpolations are visually similar but different in details
noticed by zooming out.}}
\label{fig:SR_CR}
\end{figure}

While image super-resolution tries to enhance spatial resolution from 
spatially degraded images, a related application in the
spectral domain exists, i.e., enhancing spectral resolution from a spectrally 
degraded image. One interesting example is the so-called automatic colorization,
i.e., hallucinating a plausible color version of a colorless photograph.
To the best knowledge of the authors, this is the first time we can
enhance both spectral and spatial resolutions from one single band image.
In this section, we have tested the ability to perform joint super-resolution and colorization
from one single colorless LR image on the celebA-dataset \cite{liu2015deep}.
The LR colorless image, its bicubic interpolation and $\times 2$ HR image are displayed in
the top row of Fig. \ref{fig:SR_CR}. The ADMM updates in the $1$st, $4$th
and $7$th iterations (on held-out test set) are displayed in the bottom row, showing 
that the updated image evolves towards higher quality. More results are
in the supplementary materials (see Section 3.3).


%
%

\section{Conclusion}
In this paper we have proposed an accelerated alternating direction method of multipliers, namely,
Inf-ADMM-ADNN to solve inverse problems by using two pre-trained deep neural networks.
Each ADMM update consists of feed-forward propagation through these
two networks, with a complexity of linear order with respect to the data dimensionality.
More specifically, a conditional pixel shuffling denoising auto-encoder has been learned
to perform amortized inference for the proximity operator. This auto-encoder leads to an implicit prior learned from
training data. 
A data-independent structured convolutional neural network has been learned from noise
to explicitly invert the big matrix associated with the forward model, getting rid of any
inner loop in an ADMM update, in contrast to the conventional gradient based method.
This network can also be combined with existing proximity operators to accelerate existing ADMM solvers.
Experiments and analysis on both synthetic and real dataset demonstrate
the efficiency and accuracy of the proposed method. 
In future work we hope to extend the proposed method to inverse problems 
related to nonlinear forward models.

\section*{Acknowledgments}
The authors would like to thank NVIDIA for the GPU donations.

\newpage
\begin{appendices}

\section{Learning $A$ from training data}
\begin{figure}[H]
\centering
\includegraphics[width=0.32\textwidth]{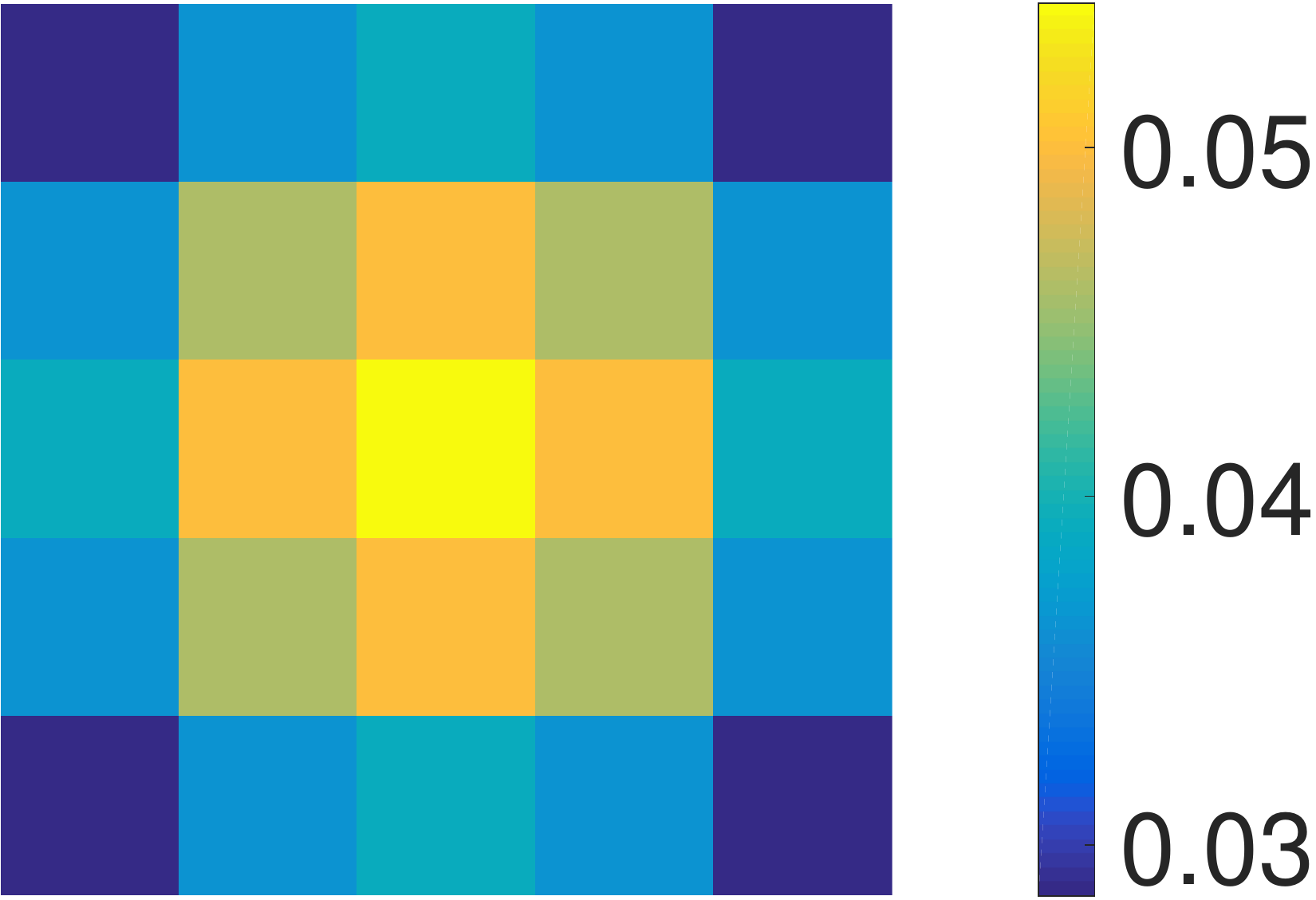}
\includegraphics[width=0.32\textwidth]{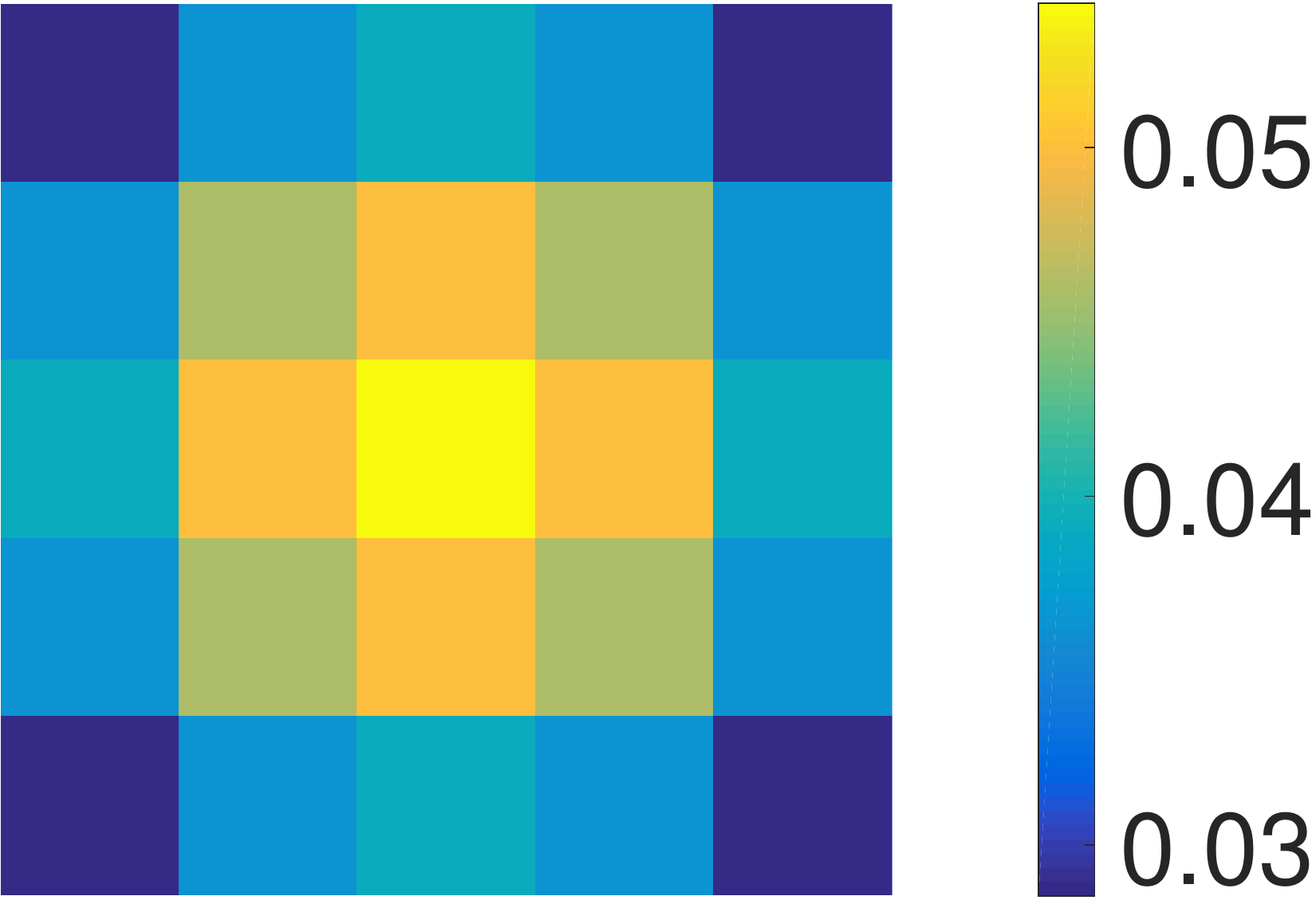}
\includegraphics[width=0.32\textwidth]{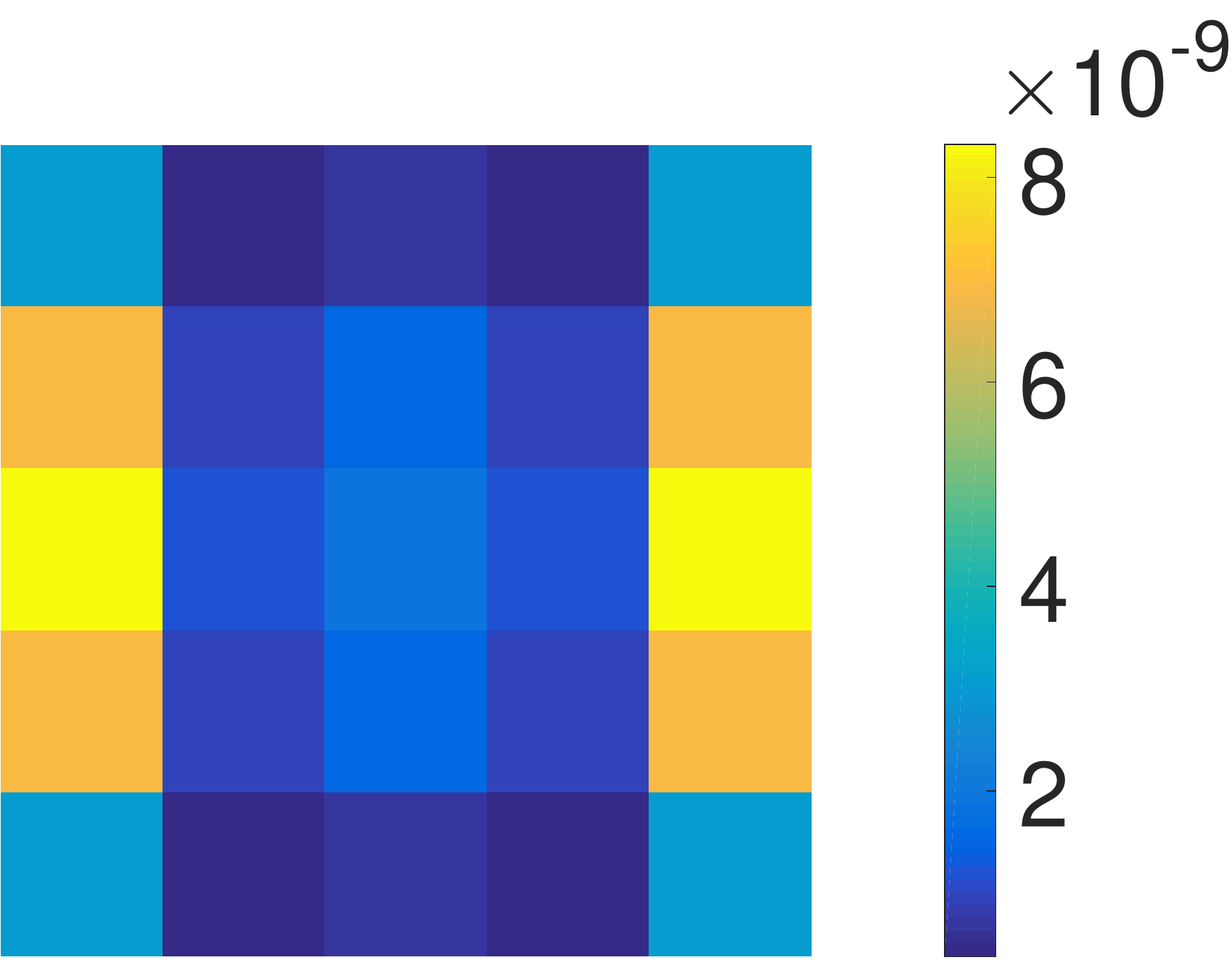}\\
\caption{(left) Ground-truth kernel for $A$ , (middle) learned kernel for $A$, (right) difference of these two.}
\label{fig:A_learn}
\end{figure}

The objective to estimate $A$ is formulated as 
\begin{equation}
\argmin_{A} \sum_{i=1}^{N}\|\bfy_i - A\bfx_i\|_2^2 + \lambda \phi(A)
\end{equation}
where $({\bfx_i, \bfy_i})_{i=1:N}$ are the training pairs and $\phi(A)$
corresponds to a regularization to $A$. Empirically, when $m$ is large enough,
the regularization plays a less important role. The learned and real kernels for $A$ (of size $4 \times 4$)
are visually very similar as is shown in Fig. \ref{fig:A_learn}.

\newpage
\begin{figure}
\includegraphics[width=0.93\textwidth]{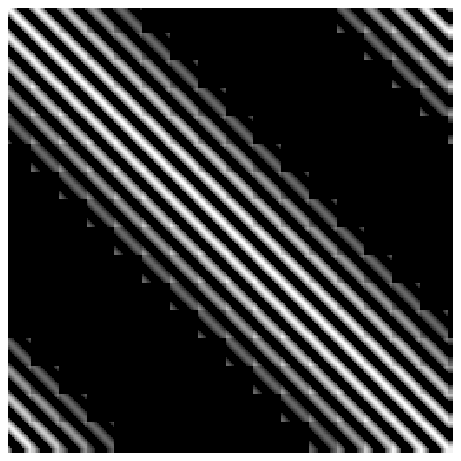} 
\includegraphics[width=0.93\textwidth]{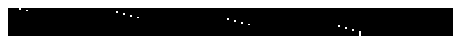}
\includegraphics[width=0.93\textwidth]{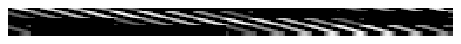}
\caption{(top) convolution matrix $H$, (middle) downsample matrix $S$,  (right) strided convolution matrix $A = SH$.}
\label{fig: HSA_2D}
\end{figure}
\section{Structure of matrix $A$ in Section 4.1}
The degradation matrix $A$ in strided convolution can be decomposed as the product of $H$ and $S$,
i.e., $A= SH$, where $H$ is a square matrix corresponding to 2-D convolution and $S$ represents the 
regular 2-D downsampling. In general, the blurring matrix $H$ is a block Toeplitz matrix
with Toeplitz blocks. If the convolution is implemented with periodic boundary conditions,
i.e., the pixels out of an image is padded with periodic extension of itself,
the matrix $H$ is a block circulant matrix with circulant blocks (BCCB).
Note that for 1-D case, the matrix $B$ reduces to a circulant matrix.
For illustration purposes, an example of matrix $B$ for a 1-D case is 
given as below.

$H=$
\begin{equation*}
\left[
			\begin{array}{ccccccccccccccccccccc}
						0.5&0.3&0&0&0&0&0&0&0&0&0&0&0&0&0&0.2\\
						0.2&0.5&0.3&0 &0&0&0&0&0&0&0&0&0&0&0&0\\
						0 &0.2&0.5&0.3&0&0&0&0&0&0&0&0&0&0&0&0\\
						0 &0 &0.2&0.5&0.3&0&0&0&0&0&0&0&0&0&0&0\\
						0 &0 &0 &0.2&0.5&0.3&0&0&0&0&0&0&0&0&0&0\\
						0 &0 &0 &0 &0.2&0.5&0.3&0&0&0&0&0&0&0&0&0\\
						0 &0 &0 &0 &0 &0.2&0.5&0.3&0&0&0&0&0&0&0&0\\
						0 &0 &0 &0 &0 &0 &0.2&0.5&0.3&0&0&0&0&0&0&0\\
						0 &0 &0 &0 &0 &0 &0 &0.2&0.5&0.3&0&0&0&0&0&0\\
						0 &0 &0 &0 &0 &0 &0 &0 &0.2&0.5&0.3&0&0&0&0&0\\
						0 &0 &0 &0 &0 &0 &0 &0 &0 &0.2&0.5&0.3&0&0&0&0\\
						0 &0 &0 &0 &0 &0 &0 &0 &0 &0 &0.2&0.5&0.3&0&0&0\\
						0 &0 &0 &0 &0 &0 &0 &0 &0 &0 &0 &0.2&0.5&0.3&0&0\\
						0 &0 &0 &0 &0 &0 &0 &0 &0 &0 &0 &0 &0.2&0.5&0.3&0\\
						0 & 0 &0 &0 &0 &0 &0 &0 &0 &0 &0 &0 &0 &0.2&0.5&0.3\\
						0.3 &0 & 0 &0 &0 &0 &0 &0 &0 &0 &0 &0 &0 &0 &0.2&0.5
			\end{array}
			\right]
\end{equation*}
An example of matrix $B$ for 2-D convolution of a $9 \times 9$ kernel with a $16 \times 16$ image 
is given in the top of Fig. \ref{fig: HSA_2D}. Clearly, in this huge matrix, 
a circulant structure is present in the block scale as well as within each block,
which clearly demonstrates the self-similar pattern of BCCB matrix.

The downsampling matrix $S$ corresponds to downsampling the original signal and its transpose
$S^T$ interpolates the decimated signal with zeros.
Similarly, a 1-D example of downsampling matrix is shown in \eqref{eq:1D_dec} for an illustrative purpose.
An example of matrix $S$ for downsampling a $16 \times 16$ image to the size 
of $4 \times 4$, i.e., $S \in \mathbb{R}^{16 \times 256}$,
is displayed in the middle of Fig. \ref{fig: HSA_2D}. The resulting degradation matrix $A$,
which is the product of $S$ and $H$ is shown in the bottom of Fig. \ref{fig: HSA_2D}.
\begin{equation}
\label{eq:1D_dec}
S=\left[
			\begin{array}{ccccccccccccccccccccc}
						1&0&0&0&{|}&0&0&0&0&{|}&0&0&0&0&{|}&0&0&0&0\\
						0&0&0&0&{|}&1&0&0&0&{|}&0&0&0&0&{|}&0&0&0&0\\
						0&0&0&0&{|}&0&0&0&0&{|}&1&0&0&0&{|}&0&0&0&0\\
						0&0&0&0&{|}&0&0&0&0&{|}&0&0&0&0&{|}&1&0&0&0
			\end{array}
			\right]
\end{equation}
\vspace{-0.3cm}
\section{More experimental results}
\subsection{Motion deblurring}
\begin{figure}[H]
\centering
\includegraphics[width=0.3\textwidth]{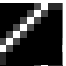}
\caption{\footnotesize{$9 \times 9$ motion blurring kernel.}}
\label{fig:mot_kernel}
\end{figure}

\begin{figure}[H]
\centering
\subfigure[ground-truth image for motion deblurring task]{
\includegraphics[width=0.8\textwidth]{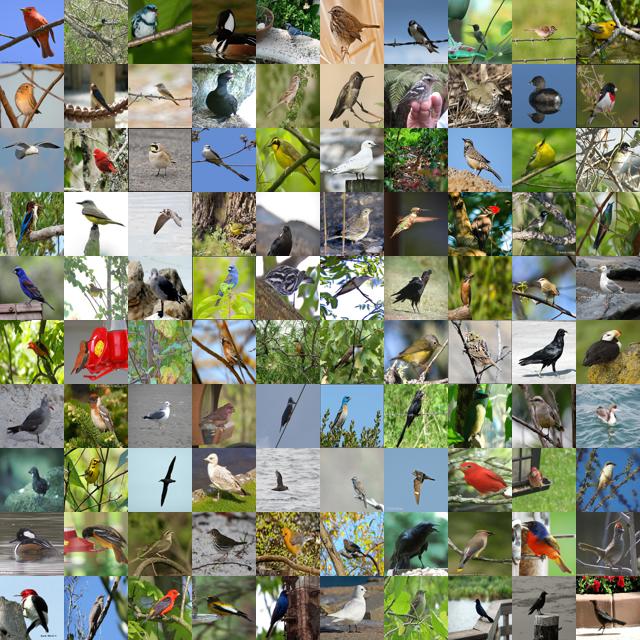}}
\subfigure[motion blurred images]{
\includegraphics[width=0.8\textwidth]{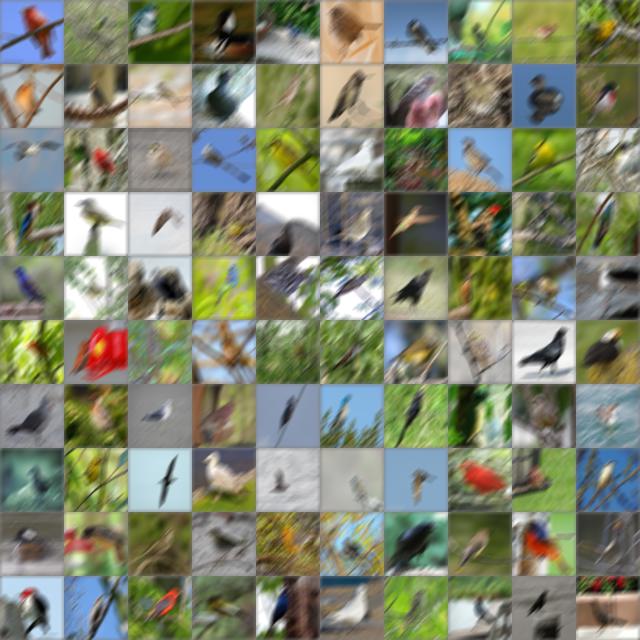}}
\caption{\footnotesize{motion blurring}}
\label{fig:mot_HR}
\end{figure}

\begin{figure}[H]
\centering
\subfigure[{results using Wiener filter with the best performance by tuning regularization parameter.}]{
\includegraphics[width=0.8\textwidth]{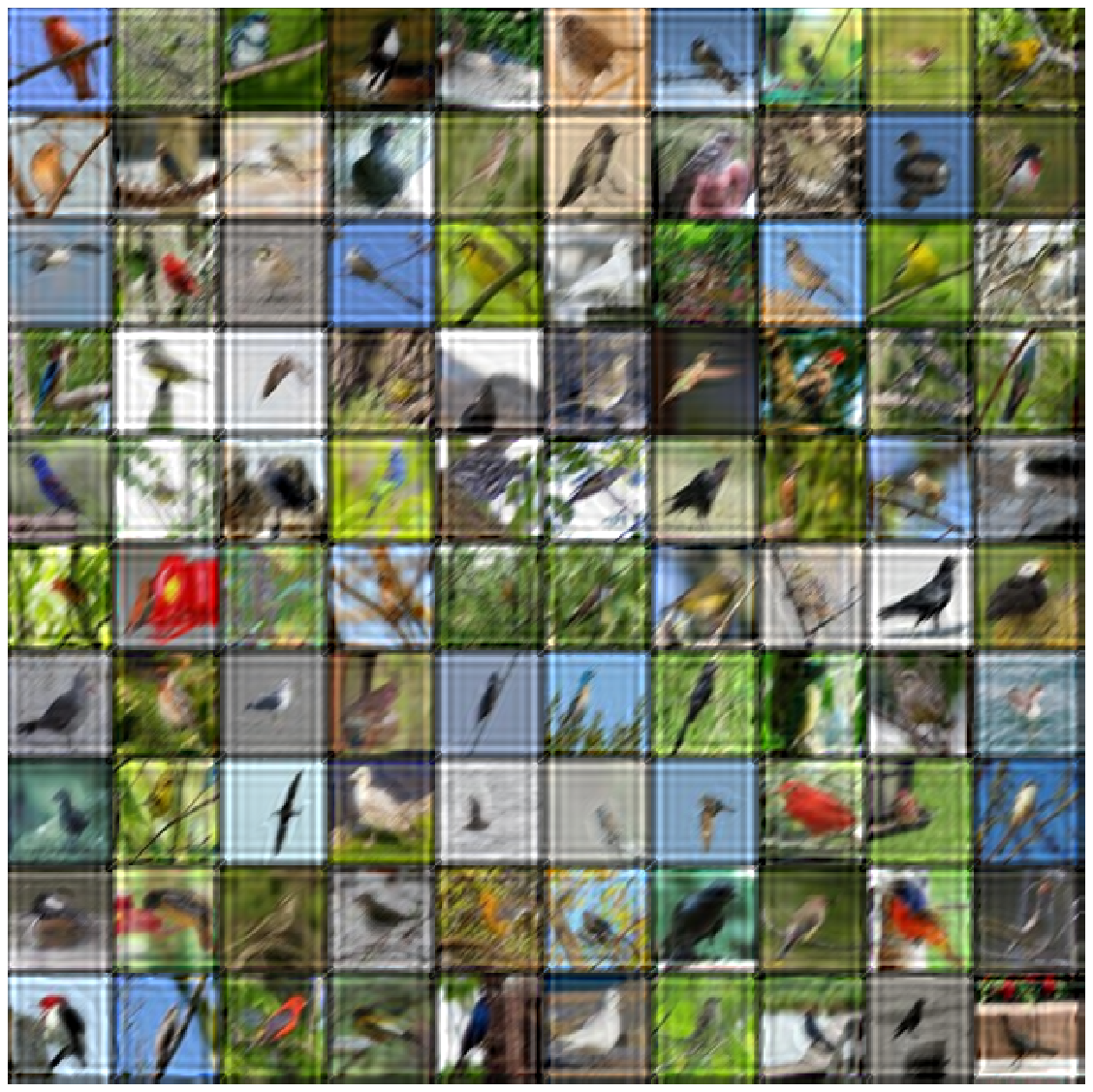}}
\subfigure[deblurred results using Inf-ADMM-ADNN]{
\includegraphics[width=0.8\textwidth]{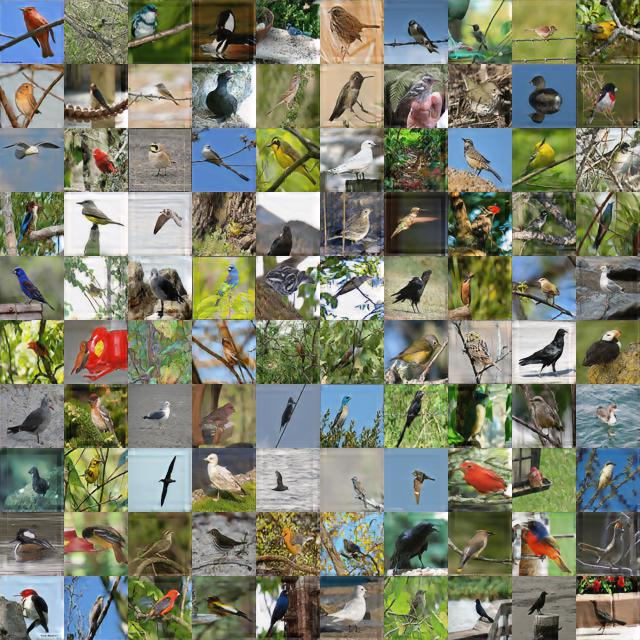}}
\caption{\footnotesize{motion deblurring}}
\label{fig:mot_LR_rec}
\end{figure}

The motion blurring kernel is shown in Fig. \ref{fig:mot_kernel}.
More results of motion deblurring on held-out testing data for CUB dataset 
are displayed in Fig. \ref{fig:mot_HR}, \ref{fig:mot_LR_rec}.
\subsection{Super-resolution}
\begin{figure}[H]
\centering
\subfigure[LR images]{
\includegraphics[width=0.8\textwidth]{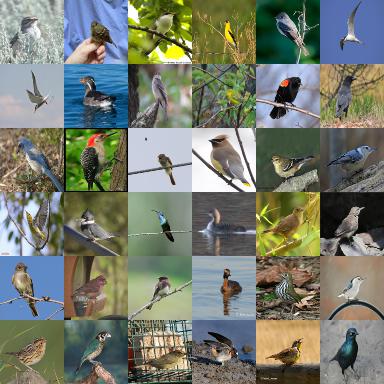}}
\subfigure[bicubic interpolations]{
\includegraphics[width=0.8\textwidth]{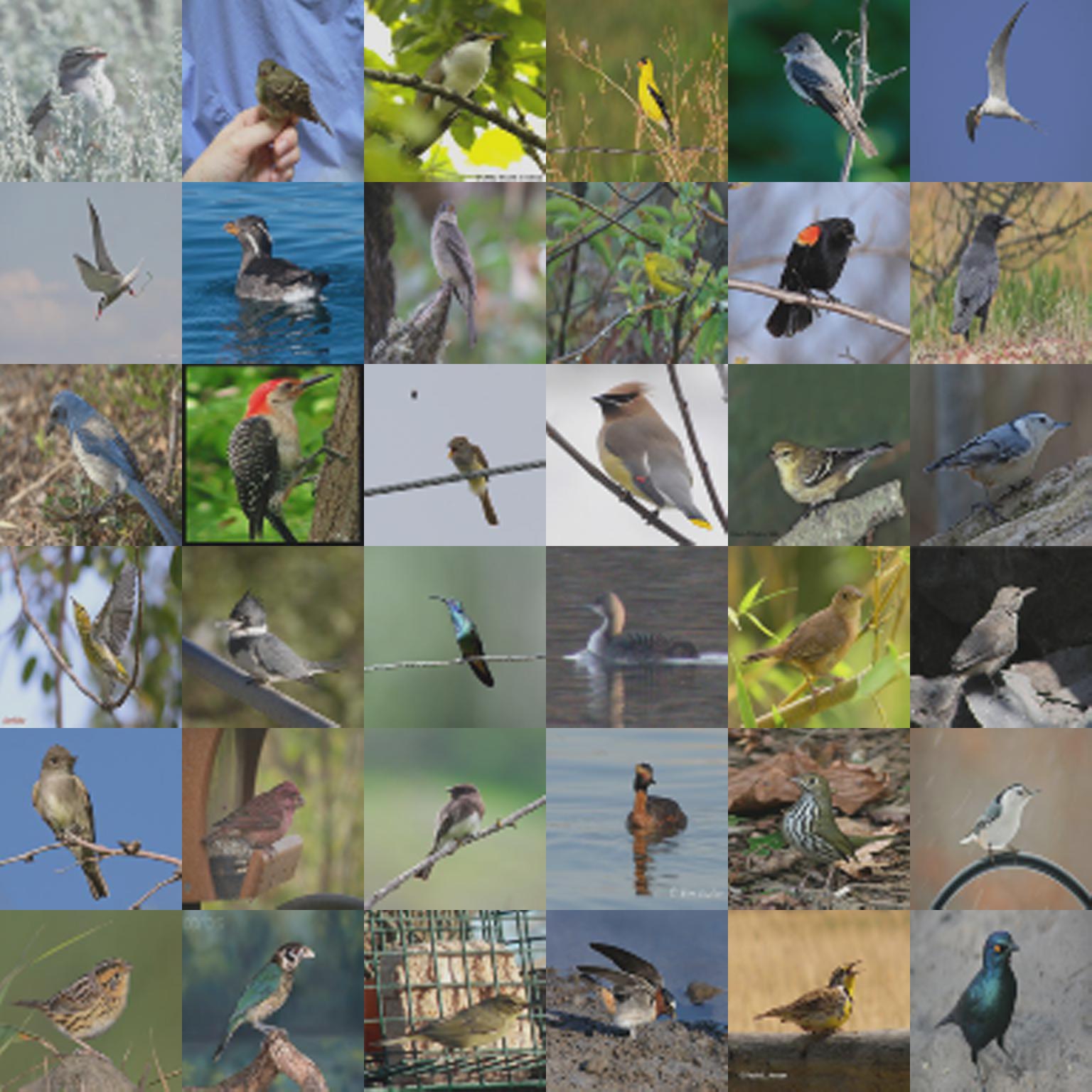}}
\caption{\footnotesize{super-resolution}}
\label{fig:SR_more_1}
\end{figure}

\begin{figure}[H]
\centering
\subfigure[super-resolved ($\times 4$)  images using Inf-ADMM-ADNN]{
\includegraphics[width=0.8\textwidth]{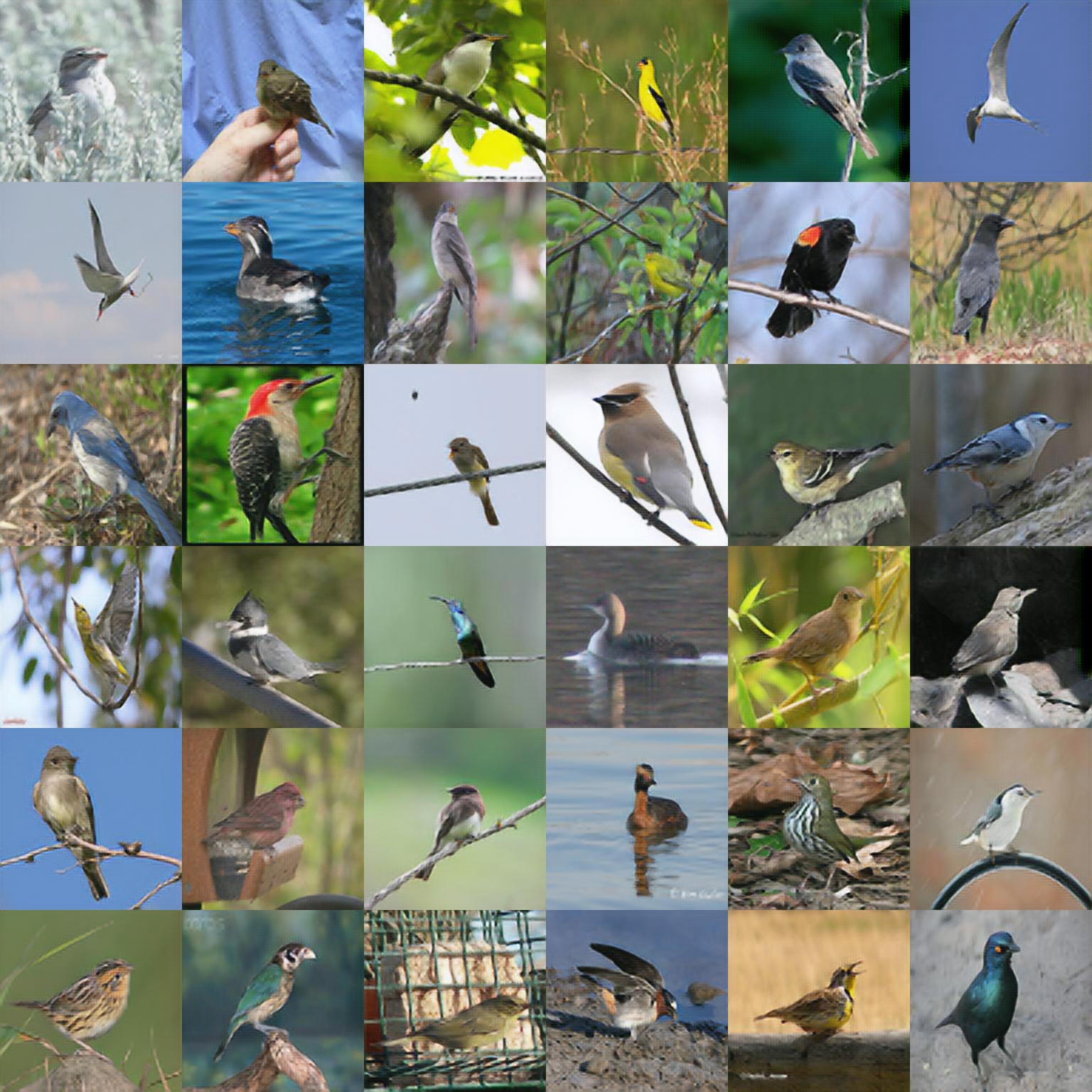}}
\subfigure[HR groundtruth]{
\includegraphics[width=0.8\textwidth]{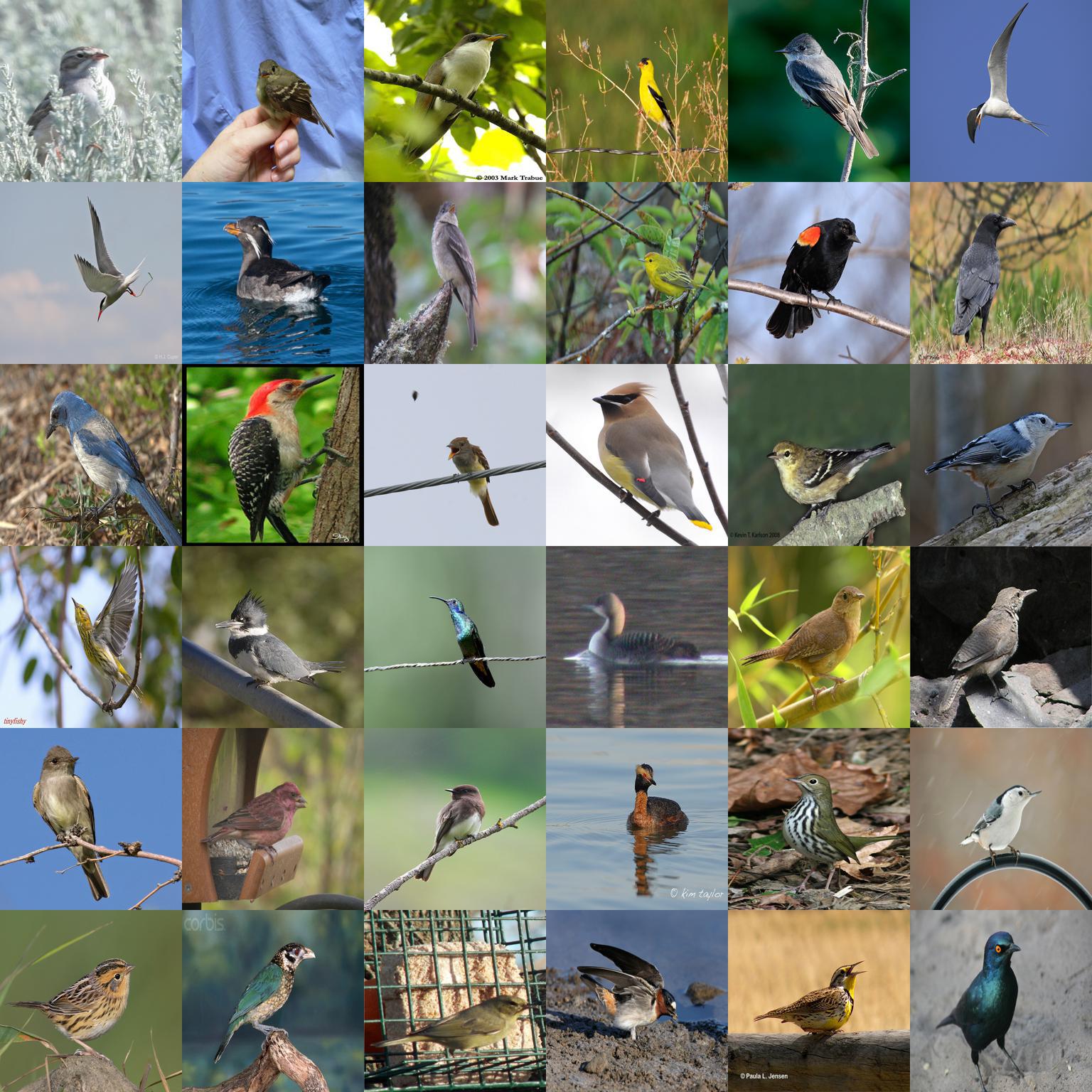}}
\caption{\footnotesize{super-resolution}}
\label{fig:SR_more_2}
\end{figure}

More results of super-resolution on held-out testing data for CUB dataset
are displayed in Fig. \ref{fig:SR_more_1}, \ref{fig:SR_more_2}.

\vspace{-0.3cm}
\subsection{Joint super-resolution and colorization}
\begin{figure}[H]
\centering
\subfigure[colorless images]{
\includegraphics[width=0.5\textwidth]{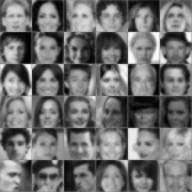}}
\subfigure[joint super-resolution ($\times 2$) and colorization using Inf-ADMM-ADNN]{
\includegraphics[width=0.5\textwidth]{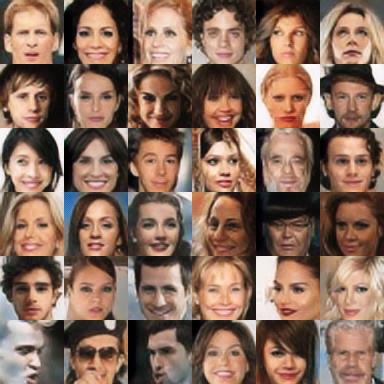}}
\subfigure[HR groundtruth]{
\includegraphics[width=0.5\textwidth]{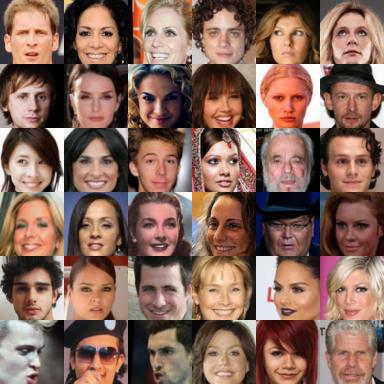}}
\caption{\footnotesize{joint super-resolution and colorization}}
\label{fig:joint_more}
\end{figure}

More results of joint super-resolution and colorization on held-out testing data 
for CelebA dataset are displayed in Fig. \ref{fig:joint_more}.

\newpage
\section{Networks Setting}
\subsection{Network for updating $\bfx$}
For MNIST dataset, we did not use the pixel shuffling strategy, since each data point is a $28 \times 28$ grayscale image,
which is relatively small. Alternatively, we used a standard denoising auto-encoder with architecture specifications 
in Table \ref{tab:mnist}.
\begin{table}[h]
\centering
\caption{Network Hyper-Parameters of DAE for MNIST}
\begin{tabular}{ c | c  | c  } 
\hline
Input Dim    & Layer                                           & Output Dim \\
\hline
$ 28 \times 28 \times 1 $ & Conv($5,5,1,32$)-Stride($2,2$)-`SAME'-Relu          & $14 \times 14 \times 32$ \\
$ 14 \times 14 \times 32 $& Conv($5,5,32,64$)-Stride($2,2$)-`SAME'-Relu         & $7 \times 7 \times 64 $\\
$ 7 \times 7 \times 64 $  & Conv($5,5,64,128$)-Stride($2,2$)-`VALID'-Relu       & $2 \times 2 \times 128 $\\
$ 2 \times 2 \times 128$  & Conv($2,2,128,64$)-Stride($1,1$)-`VALID'-None       & $1 \times 1 \times 64 $\\
$ 1 \times 1 \times 64  $ & Conv\_trans($3,3,64,128$)-Stride($1,1$)-`VALID'-Relu & $3 \times 3 \times 128$ \\
$ 3 \times3 \times 128$  & Conv\_trans($5,5,128,64$)-Stride($1,1$)-`VALID'-Relu & $7 \times 7 \times 64 $\\
$ 7 \times 7 \times 64 $  & Conv\_trans($5,5,64,32$)-Stride($2,2$)-`SAME'-Relu   & $14 \times 14 \times 32$ \\
$ 14 \times 14 \times 32 $& Conv\_trans($5,5,32,1$)-Stride($2,2$)-`SAME'-Sigmoid & $28 \times 28 \times 1 $\\
\hline
\end{tabular}\label{tab:mnist}
\end{table}

For CUB-200-2011 dataset, we applied a periodical pixel shuffling layer to the input image
of size $256 \times 256 \times 3$ with the output of size $64 \times 64 \times 48$.
Note that we did not use any stride here since we keep the image scale in each layer identical. 
The architecture of the cPSDAE is given in Table \ref{tab:cub}.
For CelebA dataset, we applied the periodical pixel shuffling layer to the input image of
size $128 \times 128 \times 3$ with the output of size $32 \times 32 \times 48$, 
and the rest of setting is the same as CUB-200-2011 dataset, as shown in Table \ref{tab:celeb}.
In terms of the discriminator, we fed the pixel shuffled images. The architecture of the 
disriminator is the same as the one in DCGAN. 

\begin{table}[h]
\centering
\caption{Network hyper-parameters of cPSDAE for CUB-200-2011}
\begin{tabular}{ |c|c|c| } 
\hline
Input Dim    & Layer                                           & Output Dim \\
\hline
$256 \times 256 \times 3$      & periodical pixel shuffling                     & $64 \times  64 \times 48$ \\
$64 \times 64 \times 48$       & Conv($4,4,48,128$)-`SAME'-Batch\_Norm-Relu       & $64 \times 64 \times 128$ \\
$64 \times 64 \times 128$      & Conv($4,4,128,64$)-`SAME'-Batch\_Norm-Relu       & $64 \times 64 \times  64 $ \\
$64 \times 64 \times 64$       & Conv($4,4,64,32$)-`SAME'-Batch\_Norm-Relu        & $64 \times 64 \times 32$ \\
$64 \times 64 \times \{32,3\}$ & Concatenate in Channel                         & $64 \times 64 \times 35 $ \\
$64 \times 64 \times 35$       & Conv($4,4,35,64$)-`SAME'-Batch\_Norm-Relu        & $64 \times 64 \times 64 $\\
$64 \times 64 \times 64 $      & Conv($4,4,64,128$)-`SAME'-Batch\_Norm-Relu       & $64 \times 64 \times 128$ \\
$64 \times 64 \times 128 $     & Conv($4,4,128,48$)-`SAME'-Batch\_Norm-Relu       & $64 \times 64 \times 48$ \\
$64 \times 64 \times 48  $     & periodical pixel shuffling                     & $256 \times 256 \times 3$ \\
\hline
\end{tabular}\label{tab:cub}
\end{table}

\begin{table}[h]
\centering
\caption{Network hyper-parameters of cPSDAE for CelebA}
\begin{tabular}{ |c|c|c| } 
\hline
Input Dim    & Layer                                           & Output Dim \\
\hline
$64 \times 64 \times 3$      & periodical pixel shuffling                     & $32 \times  32 \times 12$ \\
$32 \times 32 \times 12$       & Conv($4,4,12,128$)-`SAME'-Batch\_Norm-Relu       & $32 \times 32 \times 128$ \\
$32 \times 32 \times 128$      & Conv($4,4,128,64$)-`SAME'-Batch\_Norm-Relu       & $32 \times 32 \times  64 $ \\
$32 \times 32 \times 64$       & Conv($4,4,64,32$)-`SAME'-Batch\_Norm-Relu        & $32 \times 32 \times 32$ \\
$32 \times 32 \times \{32,3\}$ & Concatenate in Channel                         & $32 \times 32 \times 35 $ \\
$32 \times 32 \times 35$       & Conv($4,4,35,64$)-`SAME'-Batch\_Norm-Relu        & $32 \times 32 \times 64 $\\
$32 \times 32 \times 64 $      & Conv($4,4,64,128$)-`SAME'-Batch\_Norm-Relu       & $32 \times 32 \times 128$ \\
$32 \times 32 \times 128 $     & Conv($4,4,128,12$)-`SAME'-Batch\_Norm-Relu       & $32 \times 32 \times 12$ \\
$32 \times 32 \times 12  $     & periodical pixel shuffling                     & $64 \times 64 \times 3$ \\
\hline
\end{tabular}\label{tab:celeb}
\end{table}

\subsection{Network for updating $\bfz$}
As described in Section 3.2, the neural network to update $\bfz$ 
was designed to have symmetric architecture. 
The details of this architecture is given in Table \ref{tab:inverse}.
Note that $W \times H$ represents the size of the width and height of 
measurement $\bfy$.
\begin{table}[htbp]
\centering
\caption{Symmetric network hyper-parameters for updating $\bfz$}
\begin{tabular}{ |l|l|l| } 
\hline
Input Dim    & Layer  & Output Dim \\
\hline
$H$ $\times$ $W$ $\times$ 3  & Conv\_trans(4,4,3,32, $W_0$)-`SAME'-Relu        & $H$ $\times$ $W$ $\times$ 32 \\
$H$ $\times$ $W$ $\times$ 32 & Conv\_trans(4,4,32,64, $W_1$)-`SAME'-Relu       & $H$ $\times$ $W$ $\times$ 64 \\
$H$ $\times$ $W$ $\times$ 64 & Conv(4,4,3,32, $W_1$)-`SAME'-Relu               & $H$ $\times$ $W$ $\times$ 32 \\
$H$ $\times$ $W$ $\times$ 32 & Conv(4,4,32,64, $W_0$)-`SAME'                   & $H$ $\times$ $W$ $\times$ 3 \\
\hline
\end{tabular}\label{tab:inverse}
\end{table}

\end{appendices}

\newpage
\bibliographystyle{plain}
\bibliography{strings_all_ref,reference}

\begin{thebibliography}{10}

\bibitem{adler2017solving}
Jonas Adler and Ozan {\"O}ktem.
\newblock Solving ill-posed inverse problems using iterative deep neural
  networks.
\newblock {\em arXiv preprint arXiv:1704.04058}, 2017.

\bibitem{Boyd2011}
Stephen Boyd, Neal Parikh, Eric Chu, Borja Peleato, and Jonathan Eckstein.
\newblock Distributed optimization and statistical learning via the alternating
  direction method of multipliers.
\newblock {\em Foundations and Trends{\textregistered} in Machine Learning},
  3(1):1--122, 2011.

\bibitem{bruna2015super}
Joan Bruna, Pablo Sprechmann, and Yann LeCun.
\newblock Super-resolution with deep convolutional sufficient statistics.
\newblock {\em arXiv preprint arXiv:1511.05666}, 2015.

\bibitem{chang2017one}
JH~Chang, Chun-Liang Li, Barnabas Poczos, BVK Kumar, and Aswin~C
  Sankaranarayanan.
\newblock One network to solve them all---solving linear inverse problems using
  deep projection models.
\newblock {\em arXiv preprint arXiv:1703.09912}, 2017.

\bibitem{Csaji2001}
Bal{\'a}zs~Csan{\'a}d Cs{\'a}ji.
\newblock Approximation with artificial neural networks.
\newblock {\em Faculty of Sciences, Etvs Lornd University, Hungary}, 24:48,
  2001.

\bibitem{Deng2014deep}
Li~Deng, Dong Yu, et~al.
\newblock Deep learning: methods and applications.
\newblock {\em Foundations and Trends{\textregistered} in Signal Processing},
  7(3--4):197--387, 2014.

\bibitem{dosovitskiy2016generating}
Alexey Dosovitskiy and Thomas Brox.
\newblock Generating images with perceptual similarity metrics based on deep
  networks.
\newblock In {\em Advances in Neural Information Processing Systems}, pages
  658--666, 2016.

\bibitem{Elad2006}
Michael Elad and Michal Aharon.
\newblock Image denoising via sparse and redundant representations over learned
  dictionaries.
\newblock {\em IEEE Trans. Image Process.}, 15(12):3736--3745, 2006.

\bibitem{gatys2016image}
Leon~A Gatys, Alexander~S Ecker, and Matthias Bethge.
\newblock Image style transfer using convolutional neural networks.
\newblock In {\em Proc. IEEE Int. Conf. Comp. Vision and Pattern Recognition
  (CVPR)}, pages 2414--2423, 2016.

\bibitem{goodfellow2014generative}
Ian Goodfellow, Jean Pouget-Abadie, Mehdi Mirza, Bing Xu, David Warde-Farley,
  Sherjil Ozair, Aaron Courville, and Yoshua Bengio.
\newblock Generative adversarial nets.
\newblock In {\em Advances in Neural Information Processing Systems}, pages
  2672--2680, 2014.

\bibitem{krizhevsky2012imagenet}
Alex Krizhevsky, Ilya Sutskever, and Geoffrey~E Hinton.
\newblock Imagenet classification with deep convolutional neural networks.
\newblock In {\em Advances in Neural Information Processing Systems}, pages
  1097--1105, 2012.

\bibitem{larsson2016learning}
Gustav Larsson, Michael Maire, and Gregory Shakhnarovich.
\newblock Learning representations for automatic colorization.
\newblock In {\em Proc. European Conf. Comp. Vision (ECCV)}, pages 577--593.
  Springer, 2016.

\bibitem{lecun1998gradient}
Yann LeCun, L{\'e}on Bottou, Yoshua Bengio, and Patrick Haffner.
\newblock Gradient-based learning applied to document recognition.
\newblock {\em Proc. IEEE}, 86(11):2278--2324, 1998.

\bibitem{lei2015predicting}
Jimmy Lei~Ba, Kevin Swersky, Sanja Fidler, et~al.
\newblock Predicting deep zero-shot convolutional neural networks using textual
  descriptions.
\newblock In {\em Proc. IEEE Int. Conf. Comp. Vision (ICCV)}, pages 4247--4255,
  2015.

\bibitem{liu2015deep}
Ziwei Liu, Ping Luo, Xiaogang Wang, and Xiaoou Tang.
\newblock Deep learning face attributes in the wild.
\newblock In {\em Proc. IEEE Int. Conf. Comp. Vision (ICCV)}, pages 3730--3738,
  2015.

\bibitem{7879849}
Songtao Lu, Mingyi Hong, and Zhengdao Wang.
\newblock A nonconvex splitting method for symmetric nonnegative matrix
  factorization: Convergence analysis and optimality.
\newblock {\em IEEE Transactions on Signal Processing}, 65(12):3120--3135, June
  2017.

\bibitem{maurer1979first}
Helmut Maurer and Jochem Zowe.
\newblock First and second-order necessary and sufficient optimality conditions
  for infinite-dimensional programming problems.
\newblock {\em Math. Progam.}, 16(1):98--110, 1979.

\bibitem{nguyen2016plug}
Anh Nguyen, Jason Yosinski, Yoshua Bengio, Alexey Dosovitskiy, and Jeff Clune.
\newblock Plug \& play generative networks: Conditional iterative generation of
  images in latent space.
\newblock {\em arXiv preprint arXiv:1612.00005}, 2016.

\bibitem{schlemper2017deep}
Jo~Schlemper, Jose Caballero, Joseph~V Hajnal, Anthony Price, and Daniel
  Rueckert.
\newblock A deep cascade of convolutional neural networks for {MR} image
  reconstruction.
\newblock {\em arXiv preprint arXiv:1703.00555}, 2017.

\bibitem{shi2016real}
Wenzhe Shi, Jose Caballero, Ferenc Husz{\'a}r, Johannes Totz, Andrew~P Aitken,
  Rob Bishop, Daniel Rueckert, and Zehan Wang.
\newblock Real-time single image and video super-resolution using an efficient
  sub-pixel convolutional neural network.
\newblock In {\em Proc. IEEE Int. Conf. Comp. Vision and Pattern Recognition
  (CVPR)}, pages 1874--1883, 2016.

\bibitem{Simoes2015}
M.~Simoes, J.~Bioucas-Dias, L.B. Almeida, and J.~Chanussot.
\newblock A convex formulation for hyperspectral image superresolution via
  subspace-based regularization.
\newblock {\em IEEE Trans. Geosci. Remote Sens.}, 53(6):3373--3388, Jun. 2015.

\bibitem{simonyan2014very}
Karen Simonyan and Andrew Zisserman.
\newblock Very deep convolutional networks for large-scale image recognition.
\newblock {\em arXiv preprint arXiv:1409.1556}, 2014.

\bibitem{sonderby2016amortised}
Casper~Kaae S{\o}nderby, Jose Caballero, Lucas Theis, Wenzhe Shi, and Ferenc
  Husz{\'a}r.
\newblock Amortised {MAP} inference for image super-resolution.
\newblock {\em arXiv preprint arXiv:1610.04490}, 2016.

\bibitem{Tarantola2005inverse}
Albert Tarantola.
\newblock {\em Inverse problem theory and methods for model parameter
  estimation}.
\newblock SIAM, 2005.

\bibitem{Tikhonov1977}
A.N. Tikhonov and V.I.A. Arsenin.
\newblock {\em Solutions of ill-posed problems}.
\newblock Scripta series in mathematics. Winston, 1977.

\bibitem{venkatakrishnan2013plug}
Singanallur~V Venkatakrishnan, Charles~A Bouman, and Brendt Wohlberg.
\newblock Plug-and-play priors for model based reconstruction.
\newblock In {\em Proc. IEEE Global Conf. Signal and Information Processing
  (GlobalSIP)}, pages 945--948. IEEE, 2013.

\bibitem{wah2011caltech}
Catherine Wah, Steve Branson, Peter Welinder, Pietro Perona, and Serge
  Belongie.
\newblock The caltech-ucsd birds-200-2011 dataset.
\newblock 2011.

\bibitem{Wei2015JSTSP}
Q.~Wei, N.~Dobigeon, and Jean-Yves Tourneret.
\newblock Bayesian fusion of multi-band images.
\newblock {\em IEEE J. Sel. Topics Signal Process.}, 9(6):1117--1127, Sept.
  2015.

\bibitem{Wei2015FastFusion}
Qi~Wei, Nicolas Dobigeon, and Jean-Yves Tourneret.
\newblock Fast fusion of multi-band images based on solving a {S}ylvester
  equation.
\newblock {\em IEEE Trans. Image Process.}, 24(11):4109--4121, Nov. 2015.

\bibitem{Wei2016RFUSE}
Qi~Wei, Nicolas Dobigeon, Jean-Yves Tourneret, J.~M. Bioucas-Dias, and Simon
  Godsill.
\newblock {R-FUSE}: Robust fast fusion of multi-band images based on solving a
  {S}ylvester equation.
\newblock {\em IEEE Signal Process. Lett.}, 23(11):1632--1636, Nov 2016.

\bibitem{Zhao2016}
N.~Zhao, Q.~Wei, A.~Basarab, N.~Dobigeon, D.~Kouam\'e, and J.~Y. Tourneret.
\newblock Fast single image super-resolution using a new analytical solution
  for $\ell_2-\ell_2$ problems.
\newblock {\em IEEE Trans. Image Process.}, 25(8):3683--3697, Aug. 2016.

\end{thebibliography}
\end{document}